\documentclass[10pt,twocolumn,letterpaper]{article}

\usepackage[pagenumbers]{cvpr} %

\usepackage[dvipsnames]{xcolor}
\usepackage{booktabs} %
\usepackage{verbatim} %
\usepackage{tcolorbox} %
\usepackage{tabu} %

\usepackage{pifont} %
\newcommand{\cmark}{\ding{51}}
\newcommand{\xmark}{\ding{55}}

\usepackage{algpseudocode}
\usepackage{algorithm} %
\usepackage{listings} %

\newcommand{\changed}[1]{{\color{black}#1}} %

\newcommand{\dataset}{BOBSL3DT\xspace} %
\newcommand{\methodname}{HandMDM\xspace} %

\usepackage[subtle]{savetrees} %

\newcommand{\appendixref}[2]{%
    \if\sepappendix1%
    #1%
    \else%
    #2%
    \fi%
}

\def\sepappendix{0}

\definecolor{cvprblue}{rgb}{0.21,0.49,0.74}
\usepackage[pagebackref,breaklinks,colorlinks,citecolor=cvprblue]{hyperref}

\def\paperID{52} %
\def\confName{3DV\xspace}
\def\confYear{2026\xspace}

\title{Text-Driven 3D Hand Motion Generation from Sign Language Data}

\author{L\'{e}ore Bensabath$^1$ \qquad Mathis Petrovich$^{2}$ \qquad G\"{u}l Varol$^1$ \\
	\small{
		$^{1}$ LIGM, \'Ecole des Ponts, IP Paris, Univ Gustave Eiffel, CNRS} \quad 
	\small{$^{2}$ NVIDIA} \\
	{\tt\small leore.bensabath@enpc.fr}\\
	{\tt\small \url{https://imagine.enpc.fr/~leore.bensabath/HandMDM}}
}

\begin{document}

\twocolumn[{%
	\renewcommand\twocolumn[1][]{#1}%
	\maketitle
	
	\begin{center}
		\centering
		\captionsetup{type=figure,skip=0pt} %
\includegraphics[width=.95\linewidth]{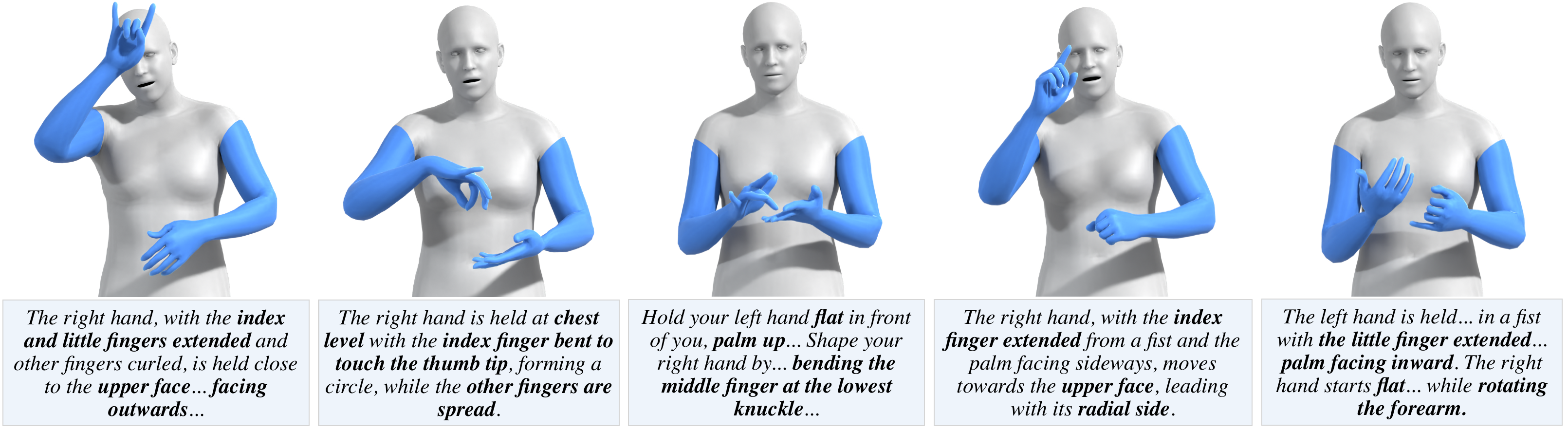}
	
\vspace{0.1cm}
\captionof{figure}{\textbf{3D hand motion from text:}
	We collect \dataset, a dataset of hand motions with accompanying textual descriptions,
	which we use to train a hand motion diffusion model (\methodname).
	We provide
	example motion-text pairs from our dataset depicting rich handshapes and movements.
	\changed{For dynamic visualizations, we refer to \appendixref{Fig.~A.6}{\cref{fig:app:teaser}} of the appendix
	and \appendixref{the supplementary video}{the video on our project page}.}
	}
\label{fig:teaser}
\end{center}
\vspace{0.1cm}
}]

\begin{abstract}
Our goal is to train a generative model of 3D hand motions, conditioned on natural language descriptions specifying motion characteristics such as handshapes, locations, finger/hand/arm movements. To this end, we automatically build pairs of 3D hand motions and their associated textual labels with unprecedented scale. Specifically, we leverage a large-scale sign language video dataset, along with noisy pseudo-annotated sign categories, which we translate into hand motion descriptions via an LLM that utilizes a dictionary of sign attributes, as well as our complementary motion-script cues. This data enables training a text-conditioned hand motion diffusion model (\methodname), that is robust across domains such as unseen sign categories from the same sign language, but also signs from another sign language and non-sign hand movements. We contribute extensive experimental investigation of these scenarios and will make our trained models and data publicly available to support future research in this relatively new field.

\end{abstract}
    
\section{Introduction}
\label{sec:intro}

Human hand motion modeling is at the center of multiple research areas, ranging from
sign languages~\cite{signavatars2024,Camgoz2020} or gestures accompanying speech~\cite{body2hands2021},
to object interactions~\cite{arctic2023}. %
In this paper, we explore the limits of \textit{text control} for generating
3D hand motions, i.e., we ask the question whether open-vocabulary language descriptions
(such as
`index and little fingers extended'
in \cref{fig:teaser})
can be used to drive a generative model of hands.
This is a relatively new task, with very recent interest in the context
of hand-object interaction~\cite{christen2024diffh2o,text2hoi2024}, and body-motion conditioning~\cite{both2hands2024}.
Here, we consider text to be the only control signal for hands, and
study complex hand motions mainly in the context of sign language data.

Recent advances in text-conditioned
human \textit{body motion} generation
have attracted a large interest~\cite{petrovich22temos,mdmtevet2023},
but one of the main limitations has been identified as the lack
of detailed \textit{hand} movements to achieve expressiveness~\cite{motionx2023}.
The reason primarily stems from lack of data.
Commonly used motion capture datasets do not contain hand articulations~\cite{amass2019};
in return, corresponding textual annotations do not describe hands~\cite{guo2022humanml3d,babel2021,kitml2016}.
There are a few exceptions with Motion-X~\cite{motionx2023} being one, where
they resort to video-based whole-body capture and
employ a rule-based script to
encode handshapes according to spatial relations among fingers, similar to
PoseScript~\cite{posescript2022} for static bodies. However, the overall database
remains still very body-focused, as also judged by the very recent work of~\cite{both2hands2024}.
To collect hand-centric annotations, \cite{both2hands2024} manually annotated the BOTH57M dataset,
capturing 3.2k
full-body motions with detailed hands in a multi-camera studio setup.
Due to the prohibitively expensive process, the resulting dataset is relatively small.

The key challenge is how to scale up the number of \textit{motions} and \textit{labels} to train a robust model.
Our solution leverages several enabling factors:
(i)~We turn to the BOBSL~\cite{albanie21bobsl} dataset, comprising 1,400 hours of British Sign Language (BSL) videos, exhaustively pseudo-labeled with sign estimates~\cite{raude24cslr2}.
(ii)~We employ state-of-the-art monocular reconstruction methods such as HAMER~\cite{hamerpavlakos2024} for hands,
and SMPLer-X~\cite{cai2023smplerx} for bodies to estimate accurate 3D upper body motions from these videos.
(iii)~We exploit SignBank~\cite{signbank2014}, a dictionary of BSL signs, with 4k \textit{glosses}\footnote{Glosses are written words associated to individual signs (\changed{see \appendixref{Appendix A}{\cref{subsec:app:glossary}}}).}, each marked with several attributes denoting handshapes, initial/final hand locations, and other phonological information. %
(iv)~We complement SignBank by detecting certain pose features from motions via rule-based scripts. %
(v)~We provide all these information per gloss category to a large language model (LLM)~\cite{gemini2024} to turn them into natural language descriptions.
With the above ingredients, we transfer textual gloss descriptions from SignBank to BOBSL sign videos by matching
the pseudo-labels to glosses, resulting in our \dataset dataset consisting of \changed{1.3M} %
motion-text pairs.

Note that there are several sources of noise, making it challenging to train a model on this data.
In particular, a given pseudo-label word may map to multiple sign variants in SignBank. Therefore, the transfer
of gloss descriptions to pseudo-labels is not straightforward, since we do not know a priori
which variant BOBSL sign corresponds to.
While we detail the exact process in \cref{sec:method}, in simple terms, we estimate the variant performed in each BOBSL sign
by matching to its closest motion from SignBank videos. 
Besides this assignment noise, the pseudo-labels themselves are not always correct, the 3D reconstructions may be imprecise,
and the LLM may not capture all the motion details.
\changed{Considering the noise-scale trade-off and following the trend in modern computer vision~\cite{radford2021clip,momeni22bsldensify}, we go towards high-scale \& high-noise regime.
}
Despite %
\changed{the sources of noise,}
we show that our dataset enables training a text-conditioned diffusion model for hand motions,
which achieves promising results across a variety of domains.
Specifically, we adapt MDM~\cite{mdmtevet2023} to generate upper body movements,
using the SMPL-X expressive body model~\cite{SMPLX2019} including hand articulations.

We demonstrate encouraging performances on a manually glossed test set of \dataset, and partition
it into \textit{seen} and \textit{unseen} signs to assess the generalization within the same sign language BSL.
We further evaluate on a different sign language, ASL, and show superior performance of our model compared to
using the relatively small training partitions of ASL-Text and MS-ZSSLR-W.
Similar conclusion is made for the non-sign dataset of BOTH57M, depicting hand articulations as part of body motions.
\changed{Our quantitative metrics primarily measure}
retrieval scores using our newly trained TMR~\cite{petrovich2023tmr} model
specialized on hand motions.

Our contributions are the following:
(i)~We present a methodology to scale up the collection of motion-text pairs for describing hand and upper body motions.
(ii)~Using the resulting dataset \dataset, we train \methodname, a hand motion diffusion model conditioned on text, and obtain promising results for this relatively new task.
(iii)~We conduct experiments to showcase the generalization performance of our model across several domains with increasing gap. %
Our code, data, and trained models will be publicly available.

\section{Related work}
\label{sec:related}

\begin{table}
    \centering    
    \setlength{\tabcolsep}{3pt}
    \resizebox{1\linewidth}{!}{
        \begin{tabular}{lrrcl}
            \toprule
            Dataset & \#motions & vocab. & hand descr. & type \\
            \midrule
            BEAT \cite{beat2022} & 2,508 & - & \xmark & speech \\
            \midrule
            SGNify \cite{forte2023} & 53 & - & \xmark & SL mocap \\
            ASL3DWord \cite{dong2024signavatar} & 1,547 & - & \xmark & SL videos \\ %
            NeuralSignActors \cite{neuralsignactors2024} & 35,000  & - & \xmark & SL videos  \\ %
            SignAvatars \cite{signavatars2024} & 70,000 & - & \xmark & SL videos \\ %
            ASL-Text \cite{bilge19zsslr} & 1,598 & 154 & \cmark & SL videos \\ %
            MS-ZSSLR-W \cite{bilge23zsslr} & 8,861 & 274 & \cmark & SL videos \\ %
            \midrule
            Motion-X \cite{motionx2023} & 81,084 & 10,224 & \cmark & mocap, body-focused \\ %
            BOTH57M \cite{both2hands2024} $\dagger$ & 1,815 & 612 & \cmark & mocap, hand-focused \\ %
            \midrule
            \dataset (Ours) & \changed{1,312,339} & \changed{1,500} & \cmark & SL videos \\
            \bottomrule
        \end{tabular}
    }
    \vspace{-0.2cm}
    \caption{\textbf{3D hand motion datasets:} We summarize datasets that contain hand motions and various types of text annotations (e.g., speech, sign categories, hand descriptions).
        Sign language (SL) datasets capturing motions in studio \cite{forte2023}
        or estimating 3D motions from videos \cite{dong2024signavatar,neuralsignactors2024,signavatars2024} do not come with hand descriptions.
        Those with hand descriptions are small \cite{bilge19zsslr,bilge23zsslr}. Full-body motion capture (mocap) datasets \cite{motionx2023,both2hands2024} are less diverse on hand motions and smaller in size than our \dataset dataset. %
        $\dagger$We report the statistics of \textit{hand} descriptions from the BOTH57M dataset, which contains
        3,229 motions from a vocabulary of 3,539 words if \textit{body} descriptions are also included. Note we calculate statistics upon downloading their data, which contain slight mismatch to the original paper.
    }
    \label{tab:related}
    \vspace{-0.3cm}
\end{table}

We briefly review relevant works on body and hand motion generation, as well as
sign language production.

\noindent\textbf{Body motion generation from text.}
Thanks to a number of text-motion datasets for full bodies, such as
KIT-ML~\cite{kitml2016},
HumanML3D~\cite{guo2022humanml3d},
BABEL~\cite{babel2021},
and Motion-X~\cite{motionx2023},
there is a growing literature on human body motion generation conditioned
on free-form language descriptions~\cite{petrovich22temos,guo2022humanml3d,tm2tguo2022,tevet2022motionclip,motiondiffusezhang2022,mdmtevet2023,guo2023momask}.
Variants of VAE~\cite{kingma2014vae} formulations were employed by initial works
that obtain promising results in this direction \cite{petrovich22temos,guo2022humanml3d}.
The recently popularized diffusion models~\cite{ho2020denoising}
have been adopted in later works such as MotionDiffuse~\cite{motiondiffusezhang2022} and MDM~\cite{mdmtevet2023}.
Much of these works generate body skeletons and perform an optimization post-processing to produce skinned
outputs such as the SMPL body model parameters~\cite{smpl2015}.
To circumvent this costly process,
Petrovich~et~al.~\cite{petrovich24stmc} incorporate SMPL representation directly in MDM training, referred to as MDM-SMPL.
In this paper, we go further by adding support for the SMPL-X~\cite{SMPLX2019} model, which enables hand motions.

\noindent\textbf{Hand motion generation.}
Works for hand generation can be grouped according to their input condition and, e.g.,
speech to gestures, object to grasps, body to hands, text to sign language translation etc.
Here, we review each category briefly.
In this paper, we particularly target conditioning \textit{only on text description} of hands, which to the best of our knowledge is a new setting.

We differentiate our work from gesture generation given speech audio and/or body ~\cite{body2hands2021,qi2023,bhattacharaya2021,ginosar2019gestures,emotiongesture2024,emog2023,zhu2023}, since
our goal is to generate given textual descriptions instructing how to move hands directly.
The focus of gesture generation is often to incorporate
emotions encoded in the audio~\cite{emotiongesture2024,emog2023,zhu2023}, or to directly infer gestures
from body motion~\cite{body2hands2021,qi2023}.

Another line of work considers object-centric hand motion.
Tasks include refining an initial hand-object configuration for better grasping~\cite{zhou2023toch,taheri2024grip},
generating a whole-body movement to reach and grasp a given object~\cite{taheri2022goal,GRAB2020}.
More related to ours, IMoS~\cite{ghosh2023imos} generates hands given an object, its position, initial body pose, and an action expressed with text
(e.g., `take picture' as text and a `camera' as the object category).
Recent work shows improvements over IMoS on this text-conditioned hand-object generation task~\cite{christen2024diffh2o,text2hoi2024,lv2024himo}.
Since these works focus on grasps, their vocabulary tends to include words such as `pick', `grab', `take',
rather than handshape or motion as in our data.

Hand motion is often studied in isolation from bodies,
and existing datasets do not annotate descriptions~\cite{interhand26m2020,renderIH2023,lin2023handdiffuse}.
More recently, there are efforts to consider hands in conjunction with body motion.
Thanks to expressive body models that involve hand articulations,
such as SMPL-X~\cite{SMPLX2019}, SMPL-H~\cite{MANO2017}, Adam~\cite{joo2018totalcapture}, GHUM~\cite{ghum2020},
whole-body datasets have been collected~\cite{both2hands2024,motionx2023}.
However, the textual descriptions of Motion-X~\cite{motionx2023} primarily focus on bodies, and lack
significant details on hands, which are automatically annotated.
BOTH57M~\cite{both2hands2024} attempts addressing this limitation by collecting
a new motion capture dataset and annotating hands, as well as bodies, with text.
But as can be seen in \cref{tab:related}, the total number of motion-text pairs are much smaller
than our \dataset. Furthermore,~\cite{both2hands2024} aims to generate hands
conditioned on the body movement, and not only on text.
On the other hand, the recent Nymeria~\cite{ma2024nymeria} collection annotates hand/arm movements, but do not capture
3D fingers beyond the wrist.

Due to the inherent presence of hands in sign language (SL),
a large category of works investigate hand motion generation given
individual sign categories, or sentence translations~\cite{signavatars2024,neuralsignactors2024,dong2024signavatar,zuo2024signstokens}.
In the following, we provide an overview of SL generation in more detail.

\noindent\textbf{Sign language generation.}
Given that until recently, available SL datasets included only 2D motion modalities,
most approaches for sign language production (SLP) derive 2D body keypoints from videos using
OpenPose~\cite{openpose2019} and lift to 3D using the method proposed by~\cite{zelinka2020neural}.
Examples include Progressive Transformers~\cite{saunders2020progressive},
and its followups~\cite{saunders2020adversarial,saunders2021mixed,hwang2021nonautoregressive}.
Several works use adversarial training to turn the keypoints into a realistic video of a signer~\cite{stoll2018sign,stoll2020textsign,saunders2020adversarial,saunders2022signing,ventura2020everybody}.
The work of~\cite{stoll2022there} introduces an additional 3D mesh regressor to obtain the SMPL-X~\cite{SMPLX2019} parameters from keypoints.

A large portion of the above prior work on SLP uses the small-scale
Phoenix datasets~\cite{phoenix_cslr_2015,phoenix_slt_2018}.
More recently, several efforts have been made to create larger SL datasets with 3D motion annotations.
Three concurrent works in 2024~\cite{neuralsignactors2024,dong2024signavatar,signavatars2024}
have estimated 3D from signing videos 
to obtain SMPL-X body model~\cite{SMPLX2019} reconstructions in different ways.
SignAvatars~\cite{signavatars2024} adopts a combination of OSX~\cite{lin2023onestage}, PARE~\cite{kocabas2021pare}, ACR~\cite{yu2023acr},
as well as a fitting step for estimating SMPL-X fits on a collection of multiple SL datasets.
NeuralSignActors \cite{neuralsignactors2024} similarly applies fitting initialized by OSX~\cite{lin2023onestage} and Mediapipe~\cite{lugaresi2019mediapipe} estimates on How2Sign videos~\cite{duarte2021howsign}.
The very recent work SOKE \cite{zuo2024signstokens} complements these data for CSL-Daily videos \cite{zhou2021csldaily}, with a similar approach, also using WiLoR~\cite{potamias2025wilor} for hand estimates.
Finally, \cite{dong2024signavatar} creates the ASL3DWord isolated sign dataset based on WLASL videos~\cite{li2020wordlevel},
leveraging the SMPL-X pose estimation model Hand4Whole~\cite{moon2022accurate}.
As shown in \cref{tab:related}, the SL video sources for these works are much smaller than BOBSL~\cite{albanie21bobsl} used in our work.
More importantly, they do not contain textual descriptions for hands, but only translations, mainly in the form of sentences.

Among SL datasets that contain motion descriptions, Bilge~et~al.~\cite{bilge19zsslr} provide
textual descriptions for a small number of ASL signs (i.e., 250)
by manually assigning ASLLVD~\cite{neidle2012asllvd} isolated sign videos with textual descriptions from Webster American Sign Language Dictionary.
In their followup work~\cite{bilge23zsslr}, a larger MS-ZSSLR is introduced, building on 200 sign categories from the MSASL~\cite{joze2019msasl} videos.
As compared in \cref{tab:related}, both the number of motions and the vocabulary of descriptions are smaller than our \dataset dataset introduced
in this work.
Moreover, their primary focus is zero-shot sign language recognition using video inputs. In this paper, we extract 3D motion estimates and
use these datasets to show our model's transfer performance.

\section{Methodology}
\label{sec:method}

\begin{figure*}
    \centering
    \includegraphics[width=.95\linewidth]{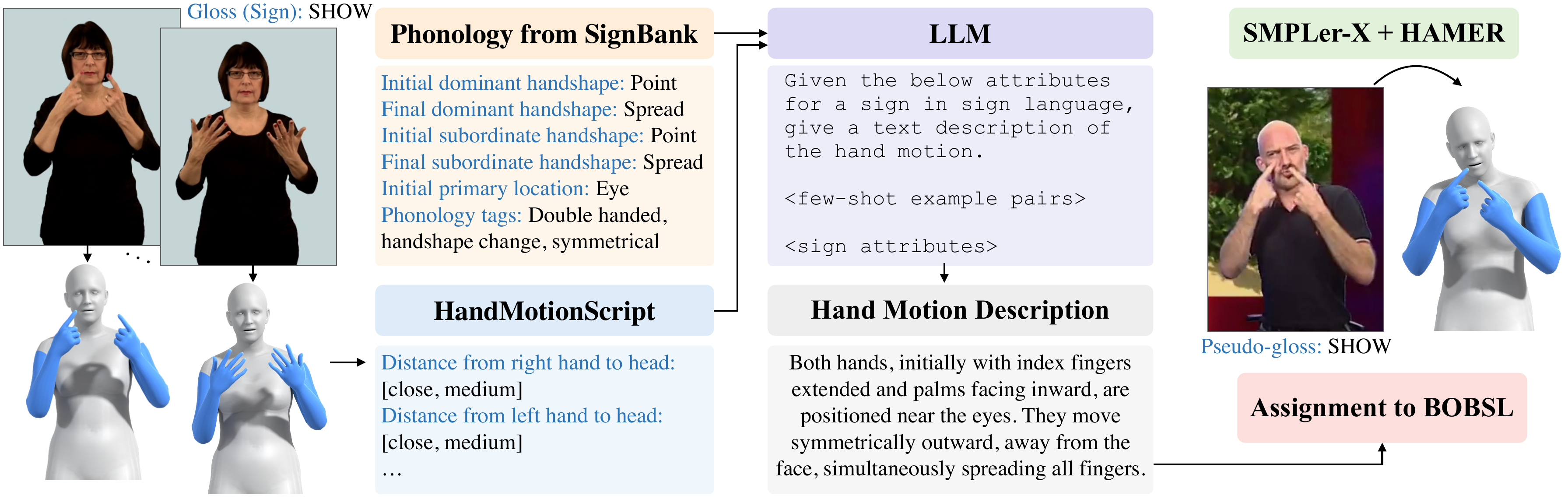}
    \vspace{-0.3cm}
    \caption{\textbf{Our approach:}
        We illustrate an overview for our data collection methodology, enabling to generate free-form textual descriptions via LLM prompting, given phonological attributes from a sign language dictionary (SignBank \cite{signbank2014}),
        as well as those detected with our HandMotionScript. The generated
        hand motion descriptions are then assigned to video-based BOBSL \cite{albanie21bobsl}
        motions thanks to the automatic pseudo-glosses combined with
        our THMR-based assignment.
        We employ a combination of SMPLer-X~\cite{cai2023smplerx} and HAMER~\cite{hamerpavlakos2024} to extract 3D motions from SignBank and BOBSL videos.
    }
    \vspace{-0.6cm}
    \label{fig:method}
\end{figure*}

We first present our methodology to automatically scale up the
motion-text pairs using sign language data (\cref{subsec:data}).
We then describe the training of text-conditioned diffusion models
to perform hand motion generation (\cref{subsec:training}).
The overall approach is illustrated in \cref{fig:method}.

\subsection{Automatic data collection} %
\label{subsec:data}

In the following, we detail the process of collecting 3D hand motions
and generating textual descriptions for them. 
We name the resulting data \dataset (videos from \textit{BOBSL}~\cite{albanie21bobsl}, with \textit{3D} estimates, \textit{T} for text).

\noindent\textbf{Source of video-sign pairs.}
In order to collect \textit{motion-text} pairs at scale, we resort to the
largest available \textit{video-sign} pairs by leveraging the BOBSL dataset~\cite{albanie21bobsl}.
The original BOBSL dataset contains 1,400 hours of continuous BSL-interpreted broadcast footage,
along with
1.2M
subtitle sentences
corresponding to the speech in the videos.
A series of works exploiting these subtitles~\cite{albanie20_bsl1k,momeni20_spotting,varol21_bslattend,momeni22bsldensify}
led to a strong isolated sign classifier~\cite{raude24cslr2}, which \changed{employs a VideoSwin~\cite{Liu2022VideoST} architecture that} takes in a short video clip (i.e., 16 frames)
and outputs the predicted sign category within a vocabulary of 8k words.
We take these \changed{frame-level} \textit{pseudo-glosses} (pseudo-labels) predicted \changed{densely} in a sliding window manner
from the entire data \changed{provided by} \cite{raude24cslr2}.
\changed{To reduce noise, we follow \cite{raude24cslr2} by filtering low-confidence frame labels and keep sign segments with at least $m=6$ repeated frames.
We additionally merge sign segments which are less than $m$ frames apart.}
This forms our basis for \changed{1.9M} \textit{video-sign} pairs across the \changed{original BOBSL} training set. 

\noindent\textbf{3D estimates from monocular videos.}
For each signing video, we follow a series of steps to obtain 3D motion estimates with accurate hands.
We adopt the state-of-the-art 3D hand reconstruction  method of HAMER~\cite{hamerpavlakos2024},
which estimates flying hands in terms of MANO \cite{MANO2017} hand pose parameters for every video frame independently.
We observe HAMER reconstructions to provide more accurate hands than other alternatives such as SMPLer-X~\cite{cai2023smplerx}.
Since, hand motions are characterized also by the rest of the upper body and especially arm movements,
we further apply the SMPLer-X expressive body reconstruction method, and stitch the HAMER hand estimates
onto these body estimates \changed{with a lightweight optimization}.
We observe that while the handshapes of
HAMER are accurate, their global locations are more stable
with SMPLer-X.
We provide \changed{more details and visual}
examples in
\appendixref{Appendix~C.3}{\cref{subsec:app:stitching}}.
By performing monocular reconstruction on video-sign pairs from BOBSL, we obtain \textit{motion-sign} pairs.

\noindent\textbf{From signs to attributes via a linguistic dictionary \& HandMotionScript.}
To associate a textual label to each sign, we propose to use a linguistic dictionary.
Specifically, we employ the phonological labeling of BSL SignBank~\cite{signbank2014},
which provides attributes about handshapes, initial/final location of each of the hands, interactions between hand and body parts,
whether a sign is one-handed or symmetric etc.
Note that these raw attributes alone do not correspond to daily vocabulary of hand motion descriptions,
but rather remain linguistic terms. 
We carefully construct a pipeline to extract textual attributes from these tags
(see \appendixref{Appendix~C.4}{\cref{subsec:app:signbank}} for further details).

While the SignBank attributes are accurate, they might be insufficient to describe the full sign, especially
lacking motion descriptions or remaining ambiguous for certain movements.
For example, the sign `SKY' involves hands moving apart from each other; however, none of the attributes refer to this dynamic motion (they denote flat handshapes etc., see \appendixref{Fig.~A.4}{\cref{fig:app:sb1}} of the appendix).
On the other hand, some of the movements can be inferred from initial/final locations, though not precisely.
To alleviate this issue, we complement the attributes with our rule-based pose descriptions, referred to
as HandMotionScript (HMS).
In particular, we implement a method based on PoseScript~\cite{posescript2022},
to detect frame-level features, 
for the hand distances to the relevant body parts and for the palm orientations.
We use SignBank attributes to determine which body parts are relevant to measure distances.
These distances are then represented as `close', `spread', etc.\ according to thresholds,
which we adapt to our fine-grained hand motion usecase. 
For example, given
a sequence of hand-to-hand distances such as [`close', `spread'],
we expect to encode the movement of `opening hands away from each other'.
We rely on the next step to convert these %
\changed{sequence of}
attributes to be interpreted as a motion.
Note that in practice, we extract %
\changed{distances and orientations}
along three axes separately,
which is described in \appendixref{Appendix~C.5}{\cref{subsec:app:hms}}.

\noindent\textbf{From attributes to textual descriptions via an LLM.}
To obtain a free-form textual description out of the above attributes, we prompt an LLM (in our case Gemini \changed{2.5 Pro}~\cite{gemini2024})
with in-context learning. We provide few-shot examples by manually designing
\changed{six} attribute-text pairs to clearly define the task for the LLM
(see \appendixref{Appendix~C.6}{\cref{subsec:app:prompt}} for the exact prompt).
With this process, we collect \textit{sign-text} pairs for 3,259 %
glosses on SignBank, corresponding to 6,221 words.
As discussed in \cref{sec:intro}, one word may correspond to multiple glosses, due to sign variants for the same concept,
or English synonyms.

For each gloss, we ask the LLM to generate three descriptions, which allows a certain level of text augmentations.
We also generate descriptions with and without HandMotionScript, resulting in six descriptions per motion.
\changed{Those without HMS are 
more hand-shape focused, while descriptions with HMS includes more dynamic information.}

We note that, in our preliminary analysis, we found that existing LLMs~\cite{gemini2024,gpt4,dubey2024llama}
do not contain knowledge of sign descriptions \changed{if only given the gloss information (without any attributes)}, which led to our exploration of SignBank and HandMotionScript.
Despite considerable effort, LLMs appear to only `know' about a small set of ASL sign descriptions,
and remain unreliable with a fair amount of hallucinations with high confidence when it comes to
describing large-vocabulary signs or non-ASL signs. For example, prompting to describe how a certain
sign performed in BSL \changed{may result} %
in description of the ASL sign.
This is why we resort to an external knowledge from a dictionary,
and also refrain from giving the gloss information in the prompt.
\changed{By going through gloss attributes instead of the gloss itself, we do not observe any concerning hallucination (see experiments in \cref{tab:input} for empirical validation).}

\noindent\textbf{Assignment of dictionary descriptions to motions.} %
Once we have a collection of motion-sign pairs (from BOBSL)
and sign-text pairs (from SignBank), the next step is the assignment of textual descriptions
into motions. This is not a simple lookup as even if the pseudo-gloss points to the correct word,
we do not have the information of which gloss variant is performed, and whether that variant is present in the SignBank dictionary (see \appendixref{Appendix~C.7}{\cref{subsec:app:assignment}} for examples).
To reduce the noise in this assignment process, we implement an automatic variant assignment in which
we look at the motion-to-motion similarity
between the 3D estimate on dictionary videos and each of the BOBSL motions.
\changed{At a high level, we select the dictionary motion among the gloss variants for a given BOBSL motion,
that maximizes the similarity. Note that in practice, we perform k-medoids clustering to remove motions that cannot be matched to any variant, and refer to \appendixref{Appendix C.7}{\cref{subsec:app:assignment}} for the exact details.}
To measure such motion-to-motion similarity,
we train a text-to-hand-motion retrieval model, \textbf{THMR}, similar to
TMR~\cite{petrovich2023tmr}, which we describe in
\appendixref{Appendix~C.2}{\cref{subsec:app:thmr}},
and use its motion encoder output as our motion embedding.
To train this model, we use random variant assignments, but observe
reliable motion-to-motion rankings since we discard the text encoder.

\changed{Our initial 1.9M video-sign pairs reduce to 1.8M \textit{motion-text} pairs after keeping the signs that have a correspondence in SignBank vocabulary. This number further drops to 1.3M after the k-medoids assignment filtering (2,351 SignBank glosses out of the initial 3,259 are assigned to BOBSL).
Finally, we remove a portion of samples to be able to build a test set of \textit{unseen} signs (see \cref{subsec:eval}) and train our model with 1.2M motion-text pairs.}

\subsection{Diffusion training with SMPL-X support} %
\label{subsec:training}

Given motion-text pairs, we train a conditional diffusion model~\cite{ho2020denoising}
to generate hand motions given text inputs.
\changed{To focus our study on the impact of automated data construction, we make minimal modifications to standard diffusion models. In particular,}
we follow MDM-SMPL \cite{petrovich24stmc} that adapts 
MDM \cite{mdmtevet2023} to have SMPL support.
We extend this approach to expressive upper body
by using SMPL-X \cite{SMPLX2019} that encodes hands and face in addition to the body pose.
We refer to this new hand motion diffusion model as \textbf{\methodname}.
Note that, while we incorporate face parameters in our training, we
did not find the SMPLer-X 3D reconstructions to be of sufficient quality
to capture the complexity of signers' facial expressions, depicting mouthing words,
or other mouth gestures and facial movements. Given that our text inputs
focus on hands,
our qualitative evaluations discard faces. %

We represent the SMPL-X parameters in 6D rotation, similar to \cite{petrovich24stmc},
and use the parameters corresponding to the upper body.
The resulting vector is of dimensionality $\mathbb{R}^{274}$ (i.e.,
15 6D joint rotations for each hand, 13 upper body 6D joint rotations, and 16 parameters for the face).
In terms of the architecture, we follow MDM-SMPL, which slightly differs from the original MDM~\cite{mdmtevet2023}, i.e.,
the inputs to the Transformer encoder are (i)~the CLIP text embedding token,
(ii)~diffusion timestep token,
(iii)~two register tokens (defined in~\cite{darcet23registers}), and (iv)~the noisy motion.
The model is trained to produce the denoised motion with a mean squared error loss.
\changed{During training, we sample a random text description among the multiple labels per motion.
We drop the condition 5\% of the time, to use classifier-free guidance at test time.
The number of diffusion steps is 100.
See \appendixref{Appendix~C.1} {\cref{subsec:app:hmdm}} for further details on the guidance coefficients, number of epochs, and other optimization details.}

\section{Experiments}
\label{sec:experiments}

In this section, we start by describing datasets and evaluation protocols used in our experiments (\cref{subsec:eval}).
We then present a series of ablation studies, such as the effect of input control, data size, and variant assignment (\cref{subsec:ablation}).
Next, we report results on out-of-domain datasets to test the transfer capabilities of our model (\cref{subsec:transfer}).
Finally, we provide qualitative analyses along with a discussion on limitations (\cref{subsec:qual}).

\subsection{Data and evaluation metrics}
\label{subsec:eval}

\noindent\textbf{\dataset-Test.}
We define a test partition of our new \dataset dataset, introduced in \cref{subsec:data}.
The work of \cite{raude24cslr2} provides manual gloss annotations for a subset of
35
BOBSL test set episodes~\cite{albanie21bobsl} \changed{spanning a vocabulary of  5.1K glosses.}
\changed{
Among these, we extract the signs that are present in SignBank and split them into seen and
unseen candidate signs (during training), especially to assess generalization with the latter. 
We further
semi-automatically construct the corresponding text descriptions
by manually verifying the SignBank gloss variant assignments.
We refer to \appendixref{Appendix~B.3}{\cref{subsec:app:testset}} for the exact process.
Our final clean test set consists of 722 seen signs and 87 unseen signs, and includes a unique motion per sign.}

\begin{table*}
    \setlength{\tabcolsep}{12pt}
    \centering    
    \resizebox{.99\linewidth}{!}{
    \begin{tabu}{l|ccc|ccc}
        \toprule
        & \multicolumn{3}{c|}{Seen} & \multicolumn{3}{c}{Unseen} \\
        Input & R@1 $\uparrow$ & R@3 $\uparrow$ & FID $\downarrow$ & R@1 $\uparrow$ & R@3 $\uparrow$ & FID $\downarrow$ \\
        \midrule
        \rowfont{\color{gray}} Gloss & 21.71$_{\pm 0.3}$ & 37.53$_{\pm 0.5}$ & 0.17 & n/a & n/a & n/a \\
        \rowfont{\color{gray}} Phonology &  17.14$_{\pm 0.8}$ & 28.78$_{\pm 0.4}$  & 0.17 & 20.98$_{\pm 3.1}$ & 34.48$_{\pm 2.2}$ & 0.40 \\
        \midrule
        LLM(Gloss) & \textcolor{white}{0}1.00$_{\pm 0.1}$ $\mid$ \textcolor{white}{0}1.00$_{\pm 0.2}$ & \textcolor{white}{0}2.29$_{\pm 0.4}$ $\mid$ \textcolor{white}{0}2.70$_{\pm 0.2}$ & 0.16 $\mid$ 0.17 & \textcolor{white}{0}5.17$_{\pm 1.3}$ $\mid$ \textcolor{white}{0}4.60$_{\pm 0.8}$ & 10.06$_{\pm 1.7}$ $\mid$ \textcolor{white}{0}8.91$_{\pm 0.5}$ & 0.56 $\mid$ 0.52  \\
        LLM(\changed{Phonology}) & 19.95$_{\pm 0.6}$ $\mid$ \textcolor{white}{0}6.30$_{\pm 0.2}$ & 33.14$_{\pm 0.9}$ $\mid$ 12.08$_{\pm 0.8}$ & \textbf{0.15} $\mid$ 0.23 & \textbf{22.99}$_{\pm 1.6}$ $\mid$ 12.64$_{\pm 2.0}$ & \textbf{37.07}$_{\pm 1.3}$ $\mid$ 21.55$_{\pm 1.0}$ & \textbf{0.38} $\mid$ 0.46  \\
        LLM(\changed{Phonology}+HMS) & \textbf{21.57}$_{\pm 0.6}$ $\mid$ \textbf{21.68}$_{\pm 0.3}$ & \textbf{35.46}$_{\pm 0.6}$ $\mid$ \textbf{34.18}$_{\pm 0.6}$ & 0.16 $\mid$ \textbf{0.16} & 21.55$_{\pm 0.5}$ $\mid$ \textbf{17.53}$_{\pm 2.2}$ & 34.19$_{\pm 1.7}$ $\mid$ \textbf{36.20}$_{\pm 2.6}$ & 0.40 $\mid$ \textbf{0.40} \\
        \bottomrule
    \end{tabu}
    }
    \vspace{-0.3cm}
    \caption{\textbf{Effect of input control:} We evaluate models trained with different input formats on the \textit{seen} and \textit{unseen} signs of the \dataset manually-glossed test set.
    There are \changed{722 / 87}
    ground-truth motion instances in these seen/unseen sets
    (\changed{with one motion instance per gloss}),
    and we report the motion-to-motion retrieval from our generation to the ground truth.
    Note that the first two rows use a different input at test time, strictly following the format used during training time (as opposed to the last \changed{three} rows that are trained and tested with free-form text). For the \changed{three} %
    variants of LLM-generated descriptions, we test on text generated only from SignBank \changed{phonology} attributes, as well as adding the HandMotionScript (HMS). We denote the two test set variants with a `$\mid$' separator, i.e., `LLM(\changed{Phonology}) $\mid$ LLM(\changed{Phonology}+HMS)'.
    }
    \label{tab:input}
    \vspace{-0.5cm}
\end{table*}

\noindent\textbf{ASL-Text \cite{bilge19zsslr}} and \textbf{MS-ZSSLR \cite{bilge23zsslr}} provide
video-text pairs for ASL signs from a vocabulary of 250 and 200 words, respectively.
ASL-Text contains 1,188 training, 151 validation, and 259 test samples, with 50 unique texts in the test set.
MS-ZSSLR has 6,107 training, 1,249 validation, 1,987 test samples, with 49 unique texts in the test set. \changed{Similarly to \dataset, we obtain the 3D motion estimates from their videos to create motion-text pairs.}
We use the training sets to train
baseline models and use the test sets for zero-shot transfer from our model.
\changed{For the zero-shot case, %
	since the test set texts include ASL-specific handshape names (28 in total), we replace these names by English text descriptions, obtained using an LLM (Gemini 2.5 Flash~\cite{gemini2024}) and manually shortening when possible.}

\noindent\textbf{BOTH57M \cite{both2hands2024}} consists of 2,583 training, 161 validation, and 485
test motions, with their corresponding textual descriptions. However, we only use a subset of 1,483/ 101/231
train/val/test samples that have \textit{hand} descriptions. We similarly evaluate the transfer performance on the test set, and use the training set to build the baseline.
\changed{Note that this is a motion capture dataset from 32 camera views. We only select from each pose the upper body joints used for \dataset.}

We use the validation set of each dataset to tune the number of epochs for baselines.

\noindent\textbf{Evaluation protocol.}
Following works reporting retrieval scores for motion generation models~\cite{petrovich24stmc,athanasiou24motionfix},
we primarily base our evaluations on retrieval metrics to account for the fact that there can be variance in motions corresponding to a given text.
We train a TMR~\cite{petrovich2023tmr} text-to-motion retrieval model, denoted as THMR, 
with the \dataset motion-text training pairs. Unlike the initial THMR model used for variant assignment (which uses random assignments, see \cref{subsec:data}),
we retrain another THMR with the automatic assignments (\appendixref{Appendix~C.2}{\cref{subsec:app:thmr}}).

For measuring performance on \dataset-Test, we report motion-to-motion retrieval scores \changed{using THMR trained using
the same representation as our upper-body motion generations}.
Specifically, given a generated motion, we measure how often the ground-truth (GT) for the corresponding text
ranks at top-k rank (R@k). \changed{We report both \textbf{R@1} and \textbf{R@3}.} 
Note that, in case of multiple GT motions per text, we mark the retrieved motion correct if the corresponding text
is \textit{exactly} the same.
For BOTH57M, we relax this constraint given the high intra-sample text similarity,
and use text-to-text similarity as in TMR~\cite{petrovich2023tmr} to determine whether a retrieval can be considered correct (i.e., if above 0.95 in CLIP text space).
We generate motions with different random noise vectors, and report the mean and the standard variance across 4 generations.

For ASL-Text~\cite{bilge19zsslr}, MS-ZSSLR~\cite{bilge23zsslr}, and BOTH57M~\cite{both2hands2024},
we observe unreliable motion-to-motion similarities, explained with the domain gap (e.g., BOTH57M motions
are much longer and noisier than \dataset). In return, the generated motions can rank irrelevant GT as high, although they correspond better to the correct GT.
We therefore resort to using motion-to-text retrieval which we find to perform relatively well. 
\changed{Given that THMR is trained on \dataset, we evaluate for these datasets only on a relevant subset of the features to avoid unfair comparison (see \appendixref{Appendix~C.2}{\cref{subsec:app:thmr}}), namely the arm and hand rotations for ASL-Text and MS-ZSSLR, and the hand rotations for BOTH57M.}

\changed{Besides the retrieval metrics, we report \textbf{FID} to measure the distance between generated and ground-truth distributions using the motion embedding space from THMR.}

\subsection{Ablation studies}
\label{subsec:ablation}

\noindent\textbf{Input control.}
We experiment with a series of input formats to control the generation.
In \cref{tab:input}, we report our results of training and testing on \dataset, reporting
seen and unseen partitions of the test set.
We first train a gloss-to-motion diffusion model, where we represent the gloss word
in its text form. While the performance on the seen test set is promising, this model
does not go beyond its pre-defined set of training glosses at test time; therefore, does not
generalize to unseen glosses, nor accepts free-form descriptions beyond signs.

Next, we 
\changed{investigate whether the information given
to the LLM to translate attributes into free-form text can be given on its own to a motion generation model. We denote this experiment by the `Phonology' row, where we train \methodname
with the text used for prompting the LLM, which contains phonological attributes detected from SignBank. While we also change the test descriptions at test time, making the rows not perfectly comparable,
we observe a drop in retrieval of the ground-truth motions
(e.g., 28.78 vs 37.53 R@3).
Besides, this model can not be applied to free-form text input, and does not easily generalize to other sign language datasets as it requires phonology attributes in SignBank format.}

\changed{We now evaluate the models trained with free-form text in the bottom block of \cref{tab:input}.
The `LLM(Gloss)' row experiments with
generating the training text descriptions by providing the LLM with only gloss names instead of attributes.
The results confirm
that the LLM does not know BSL signs, as discussed in \cref{subsec:data}, and justifies our need for the use the SignBank attributes and HandMotionScript. 
Note that the definition of `seen' does not apply to this row, because we use the clean descriptions of \dataset-Test for evaluation (those generated by giving LLM the attributes).}

\changed{The last two rows} %
of \cref{tab:input}
\changed{show our final models trained with attribute-based LLM descriptions.
Here, we also investigate}
the gain of incorporating
HandMotionScript (HMS) in the prompt, besides the SignBank \changed{phonology}. %
To obtain comparable performances, we also report on two types of test sets
that accept free-form inputs. More specifically, we analyze results
on the test set generated with %
\changed{phonology}-only, and with HMS included (separated by `$\mid$'
in the table).
Unsurprisingly, training without HMS does not work well when tested with HMS (e.g., \changed{6.30 %
R@1 seen signs}), since the model is not used to the input vocabulary also including palm orientations etc.
However, the last row is more robust across test sets due to randomly sampling among text descriptions that are %
generated both with and without HMS, \changed{demonstrating an overall improvement.}

\begin{figure}
	\includegraphics[width=1\linewidth]{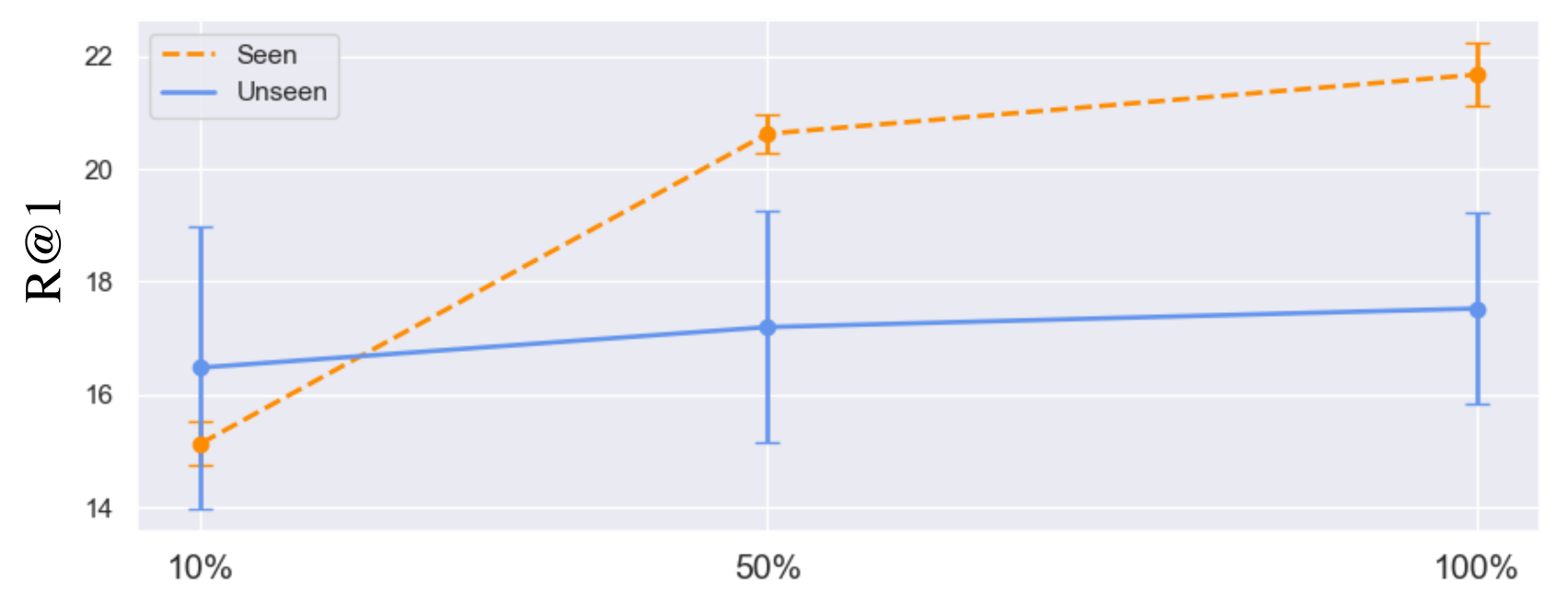}
	\vspace{-0.7cm}
	\caption{\textbf{Training data size:}
		We plot R@1 performance on the \changed{seen/unseen} test set\changed{s} of \dataset against different proportions of the training data,
		and observe a monotonic increase with larger training. 
	}
	\vspace{-0.3cm}
	\label{fig:size}
\end{figure}

\noindent\textbf{Training data size.}
In \cref{fig:size}, we analyze the impact of our large-scale training data.
We train our final \methodname model with portions of \dataset training,
and report R@1 results on \changed{seen and} unseen signs \changed{with text generated from SignBank attributes and HMS}.
We observe a steady increase with more training data,
which suggests there is room for improvement with techniques for scaling up.

\begin{table}
    \setlength{\tabcolsep}{2pt}
    \centering    
    \resizebox{1\linewidth}{!}{
    \begin{tabular}{l|ccc|ccc}
        \toprule
        & \multicolumn{3}{c|}{Seen} & \multicolumn{3}{c}{Unseen} \\
         Assignment &  R@1 $\uparrow$ & R@3 $\uparrow$ & FID $\downarrow$ & R@1 $\uparrow$ & R@3 $\uparrow$ & FID $\downarrow$ \\
        \midrule
        Random & 18.77$_{\pm 0.6}$ & 32.27$_{\pm 0.2}$ & 0.17 & 16.09$_{\pm 1.4}$ & 25.00$_{\pm 0.5}$ & 0.44 \\
        THMR & \textbf{21.68}$_{\pm 0.3}$ & \textbf{34.18}$_{\pm 0.6}$ & \textbf{0.16} & \textbf{17.53}$_{\pm 2.2}$ & \textbf{36.20}$_{\pm 2.6}$ & \textbf{0.40} \\
        \bottomrule
    \end{tabular}
    }
    \vspace{-0.3cm}
    \caption{\textbf{Assignment:}
        We observe gains with our automatic sign variant assignment
        using THMR, as opposed to randomly assigning a variant during training.
    }
    \label{tab:assignment}
    \vspace{-0.5cm}
\end{table}

\noindent\textbf{Sign variant assignment.}
As explained in \cref{subsec:data}, we use our THMR model for performing
a text assignment in case of multiple sign variants in SignBank.
In \cref{tab:assignment}, we measure the impact of this assignment,
by comparing to randomly assigning during \methodname training.
We observe a considerable improvement by reducing the noise in training by this assignment step.

\subsection{Transfer to other datasets}
\label{subsec:transfer}

We summarize our quantitative metrics for zero-shot transfer from our \dataset-trained model
to other datasets in \cref{tab:asl} and \cref{tab:both2hands}.
First, we look at a domain gap between different sign languages, BSL and ASL.
The bottom block of \cref{tab:asl} reports two variants of our model on ASL-Text and MS-ZSSLR,
where we observe that adding HandMotionScript again leads to performance boosts
by complementing the textual descriptions with motion information.
The first block of the table shows the results of training only on ASL-Text or
only on MS-ZSSLR training sets. These baselines appear to have acceptable metrics; however,
as
\changed{can be seen qualitatively}
we observe %
motions that have %
poorer fidelity
to text, while also being unrealistic \changed{potentially due to small training size}.

We make similar conclusions on the BOTH57M dataset in \cref{tab:both2hands},
which demonstrates a bigger domain gap from sign language motions.
Nevertheless, our model shows promising generalization capability to these
very different hand %
descriptions.
We note that, here, the model trained without HandMotionScript performs better, which we explain by the lack of
comprehensive motion
descriptions in BOTH57M, which tend to be more handshape focused characterized by \textit{fingers}, not as much by
arm motion or global hand location (the purpose of HMS). %

\begin{figure*}
	\centering
	\includegraphics[width=1\linewidth]{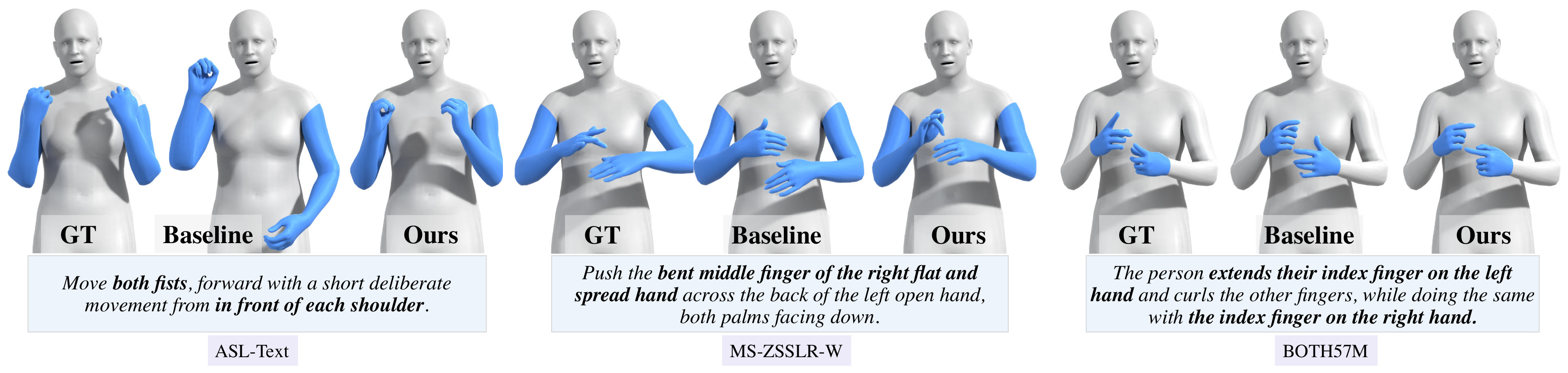}
	\vspace{-0.7cm}
	\caption{\textbf{Qualitative results for transfer:}
		We provide comparative results on the test sets of
		ASL-Text, MS-ZSSLR-W, and BOTH57M datasets.
		For each example, we show the input textual description, the ground truth (GT), the generation from the model trained from scratch on the source dataset (Baseline), and the generation from our \dataset-trained model (Ours).
		\changed{We observe limited performance of the baselines compared to ours.}
		\changed{See \appendixref{Fig.~A.7}{\cref{fig:app:qual_transfer}} of the appendix for the dynamic visualization.}%
	}
	\vspace{-0.4cm}
	\label{fig:qual_transfer}
\end{figure*}

We provide qualitative comparisons to baselines in \cref{fig:qual_transfer} for all three datasets,
and refer to \appendixref{our supplementary video}{\changed{the video on our project page}} for dynamic visuals.

\begin{table}
    \setlength{\tabcolsep}{2pt}
    \centering
    \resizebox{1\linewidth}{!}{
    \begin{tabular}{l | ccc | ccc}
    	\toprule
         & \multicolumn{3}{c|}{ASL-Text} &  \multicolumn{3}{c}{MS-ZSSLR-W}  \\
        Training data & R@1 $\uparrow$ & R@3 $\uparrow$ & FID $\downarrow$ & R@1 $\uparrow$ & R@3 $\uparrow$ & FID $\downarrow$ \\
        \midrule
        ASL-Text & \textcolor{white}{0}5.98$_{\pm 0.4}$ & 16.21$_{\pm 0.9}$ & \textbf{0.36} & 
                   \textcolor{white}{0}6.43$_{\pm 0.1}$ & 16.85$_{\pm 0.2}$ & 0.43  \\
        MS-ZSSLR-W &  \textcolor{white}{0}6.47$_{\pm 1.4}$ & 16.31$_{\pm 1.0}$ & 0.43 & 
                     \textcolor{white}{0}5.01$_{\pm 0.3}$ & 13.08$_{\pm 0.3}$ & \textbf{0.13}  \\
        \midrule
        \dataset (\changed{Phon.}) & 14.29$_{\pm 0.8}$ & 27.51$_{\pm 0.7}$ & 0.46 & 
                           12.41$_{\pm 0.4}$ & 24.52$_{\pm 0.1}$ & 0.45 \\
        \dataset (\changed{Phon.}+HMS) & \textbf{17.09}$_{\pm 1.0}$ & \textbf{35.43}$_{\pm 1.2}$ & 0.47 & 
                               \textbf{15.39}$_{\pm 0.3}$ & \textbf{31.50}$_{\pm 0.2}$ & 0.46 \\
        \bottomrule
    \end{tabular}
    }
    \vspace{-0.3cm}
    \caption{\textbf{Transfer to ASL datasets:}
    	In the top block, we show baseline performances of training \methodname from scratch on the training set of the ASL-Text or MS-ZSSLR-W dataset. Below, we report zero-shot transfer performance of our models trained on \dataset, either with LLM-generated text from SignBank \changed{phonology (Phon.),} %
    	or also by adding HandMotionScript (HMS). We report motion-to-text retrieval from generated motions to ground-truth descriptions. We see superior overall performance of our model over baselines. \changed{FID is lower when the same data source is used for training and testing due to the dataset-specific motion style.}
    }
    \vspace{-0.3cm}
    \label{tab:asl}
\end{table}

\begin{table}
    \setlength{\tabcolsep}{8pt}
    \centering
    \resizebox{.9\linewidth}{!}{
        \begin{tabular}{l | ccc}
            \toprule
            & \multicolumn{2}{c}{BOTH57M}  \\
            Training data & R@1 $\uparrow$ & R@3 $\uparrow$ & FID $\downarrow$ \\
            \midrule
            BOTH57M & 15.70$_{\pm 1.01}$ & 35.47$_{\pm 4.47}$ & \textbf{0.68} \\
                \midrule
            \dataset (\changed{Phon.}) & 16.28$_{\pm 1.64}$ & \textbf{44.19}$_{\pm 4.66}$ & 1.01 \\
            \dataset (\changed{Phon.}+HMS) & \textbf{18.61}$_{\pm 2.85}$ & 40.70$_{\pm 6.26}$ & 1.03 \\
            \bottomrule
        \end{tabular}
    }
    \vspace{-0.3cm}
    \caption{\textbf{Transfer to BOTH57M:}
        Similar to \cref{tab:asl}, we evaluate the zero-shot transfer capability of our \dataset-trained model
        by testing on BOTH57M test set of 231 motion-text pairs.
        }
    \vspace{-0.5cm}
    \label{tab:both2hands}
\end{table}

\begin{figure*}
	\includegraphics[width=1\linewidth]{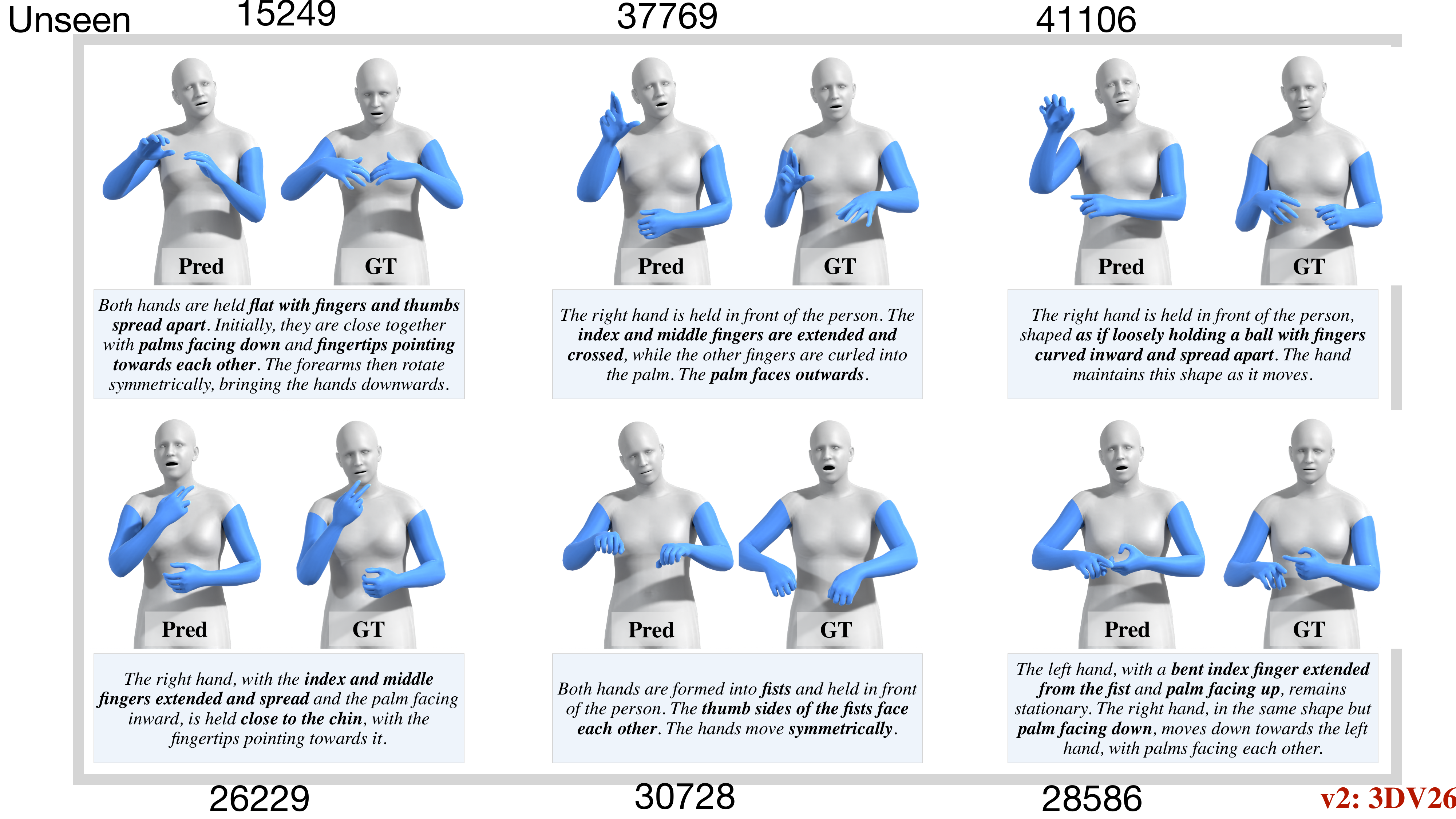}
	\vspace{-0.7cm}
	\caption{\textbf{Qualitative results \changed{on \dataset-Test}:}
		We show examples on the unseen partition of the \dataset test set.
		For each example, we display the input text, the generation (Pred) from our
		model, as well as the ground-truth (GT). Note that we visualize one static frame
		from the motions to reduce clutter, and provide a \appendixref{supplementary video}{\changed{video on our project page}},
		\changed{as well as the dynamic visualization in  \appendixref{Fig.~A.8}{\cref{fig:app:qual}} of the appendix} for further analysis.
	}
	\vspace{-0.3cm}
	\label{fig:qual}
\end{figure*}

\subsection{Qualitative analysis and limitations}
\label{subsec:qual}

In \cref{fig:qual}, we provide qualitative results for our model tested on unseen \dataset descriptions \changed{(see \appendixref{Fig.~A.9}{\cref{fig:app:qualseen}} of the appendix for results on the seen descriptions).}
Overall, we see fidelity of our generations
whenever a handshape is mentioned in the text, e.g., fist, flat, index-middle fingers extended,
all fingers spread.
Motions such as
\changed{`moves down towards the left hand' (bottom right),}
or `hands move symmetrically' (bottom middle)
are also \changed{visible in the video visualization of our predictions}. %
However, not all %
descriptions are \changed{precisely} followed
(e.g., \changed{`fists' in bottom middle do not bend the fingers sufficiently}),
especially when the 
\changed{handshape/}movement is fine-grained, or when the text contains too many details.
We also note a \changed{frequent} %
limitation of our model, not performing contact and touch movements very precisely.
\changed{For example, the fingers are not as close to the chin in the bottom left motion.}
This can be explained by the lack of precision in monocular reconstruction methods, especially
struggling with the depth dimension, leading to imperfect GT for supervision. We expect
this to be partially addressed as future reconstruction approaches improve.
An interesting avenue for future research is to explicitly detect contact
from the attributes and to enforce it in the reconstructions.

\section{Conclusion}
\label{sec:conclusion}

We described a methodology to scale up motion-text pairs for supervising text-conditioned 3D hand motion generation models.
\changed{The resulting \dataset dataset represents the first of its kind, providing over 1M text-motion pairs focusing on hands.}
Our results demonstrate that, while noisy, LLM-generated textual descriptions guided by sign language dictionary attributes
constitute a viable alternative to manually annotating limited amount of data.
\changed{This proof of concept can be extended to other sign language dictionaries to further scale up.}
We expect our model to open up a new avenue for future research in fine-grained hand movement modeling with language.

{ \small
\noindent\textbf{Acknowledgements.}
This work was granted access to the HPC resources of IDRIS under the allocations 2025-AD011014891R1, 2024-AD011013395R2 and 2025-AD011012269R4 
made by GENCI.
The authors acknowledge the ANR project CorVis ANR-21-CE23-0003-01,
and the Google Cloud Platform Credit Award enabling Gemini use.
We thank the BSL SignBank team for researcher access,
Yusuf Can Bilge for the guidance on the ASL datasets,
and Sonat Baltaci, Simon Lepage, Haran Raajesh %
for their feedback.
}

{
    \small
    \bibliographystyle{ieeenat_fullname}
    \bibliography{references}

\begin{thebibliography}{95}
\providecommand{\natexlab}[1]{#1}
\providecommand{\url}[1]{\texttt{#1}}
\expandafter\ifx\csname urlstyle\endcsname\relax
  \providecommand{\doi}[1]{doi: #1}\else
  \providecommand{\doi}{doi: \begingroup \urlstyle{rm}\Url}\fi

\bibitem[Albanie et~al.(2020)Albanie, Varol, Momeni, Afouras, Chung, Fox, and Zisserman]{albanie20_bsl1k}
Samuel Albanie, G{\"u}l Varol, Liliane Momeni, Triantafyllos Afouras, Joon~Son Chung, Neil Fox, and Andrew Zisserman.
\newblock {BSL-1K}: {S}caling up co-articulated sign language recognition using mouthing cues.
\newblock In \emph{ECCV}, 2020.

\bibitem[Albanie et~al.(2021)Albanie, Varol, Momeni, Bull, Afouras, Chowdhury, Fox, Woll, Cooper, McParland, and Zisserman]{albanie21bobsl}
Samuel Albanie, G{\"u}l Varol, Liliane Momeni, Hannah Bull, Triantafyllos Afouras, Himel Chowdhury, Neil Fox, Bencie Woll, Rob Cooper, Andrew McParland, and Andrew Zisserman.
\newblock {BOBSL}: {BBC}-{O}xford {B}ritish {S}ign {L}anguage dataset.
\newblock \emph{arXiv}, 2021.

\bibitem[Athanasiou et~al.(2024)Athanasiou, Cseke, Diomataris, Black, and Varol]{athanasiou24motionfix}
Nikos Athanasiou, Alp\'{a}r Cseke, Markos Diomataris, Michael~J. Black, and G{\"u}l Varol.
\newblock {MotionFix}: Text-driven {3D} human motion editing.
\newblock In \emph{SIGGRAPH Asia}, 2024.

\bibitem[Baltatzis et~al.(2024)Baltatzis, Potamias, Ververas, Sun, Deng, and Zafeiriou]{neuralsignactors2024}
Vasileios Baltatzis, Rolandos~Alexandros Potamias, Evangelos Ververas, Guanxiong Sun, Jiankang Deng, and Stefanos Zafeiriou.
\newblock Neural sign actors: A diffusion model for {3D} sign language production from text.
\newblock In \emph{CVPR}, 2024.

\bibitem[Bhattacharya et~al.(2021)Bhattacharya, Childs, Rewkowski, and Manocha]{bhattacharaya2021}
Uttaran Bhattacharya, Elizabeth Childs, Nicholas Rewkowski, and Dinesh Manocha.
\newblock {Speech2AffectiveGestures}: Synthesizing co-speech gestures with generative adversarial affective expression learning.
\newblock In \emph{ACM International Conference on Multimedia}, 2021.

\bibitem[Bilge et~al.(2023)Bilge, Cinbis, and Ikizler-Cinbis]{bilge23zsslr}
Y. Bilge, R. Cinbis, and N. Ikizler-Cinbis.
\newblock Towards zero-shot sign language recognition.
\newblock \emph{TPAMI}, 2023.

\bibitem[Bilge et~al.(2019)Bilge, Ikizler-Cinbis, and Cinbis]{bilge19zsslr}
Yunus~Can Bilge, Nazli Ikizler-Cinbis, and Ramazan~Gokberk Cinbis.
\newblock Zero-shot sign language recognition: Can textual data uncover sign languages?
\newblock In \emph{BMVC}, 2019.

\bibitem[Brien(1992)]{brien1992bsldict}
David Brien.
\newblock \emph{Dictionary of {British Sign Language} / {E}nglish; compiled for the {British Deaf Association} by the {Deaf Studies Research Unit}, {University of Durham}}.
\newblock Faber, 1992.

\bibitem[Cai et~al.(2023)Cai, Yin, Zeng, Wei, Sun, Yanjun, Pang, Mei, Zhang, Zhang, Loy, Yang, and Liu]{cai2023smplerx}
Zhongang Cai, Wanqi Yin, Ailing Zeng, Chen Wei, Qingping Sun, Wang Yanjun, Hui~En Pang, Haiyi Mei, Mingyuan Zhang, Lei Zhang, Chen~Change Loy, Lei Yang, and Ziwei Liu.
\newblock {SMPLer-X}: Scaling up expressive human pose and shape estimation.
\newblock In \emph{NeurIPS}, 2023.

\bibitem[Camgoz et~al.(2018)Camgoz, Hadfield, Koller, Ney, and Bowden]{phoenix_slt_2018}
Necati~Cihan Camgoz, Simon Hadfield, Oscar Koller, Hermann Ney, and Richard Bowden.
\newblock Neural sign language translation.
\newblock In \emph{CVPR}, 2018.

\bibitem[Camgoz et~al.(2020)Camgoz, Koller, Hadfield, and Bowden]{Camgoz2020}
Necati~Cihan Camgoz, Oscar Koller, Simon Hadfield, and Richard Bowden.
\newblock Multi-channel transformers for multi-articulatory sign language translation.
\newblock In \emph{ECCVW}, 2020.

\bibitem[{Cao} et~al.(2019){Cao}, {Hidalgo Martinez}, {Simon}, {Wei}, and {Sheikh}]{openpose2019}
Z. {Cao}, G. {Hidalgo Martinez}, T. {Simon}, S. {Wei}, and Y.~A. {Sheikh}.
\newblock Openpose: Realtime multi-person {2D} pose estimation using part affinity fields.
\newblock \emph{TPAMI}, 2019.

\bibitem[Cha et~al.(2024)Cha, Kim, Yoon, and Baek]{text2hoi2024}
Junuk Cha, Jihyeon Kim, Jae~Shin Yoon, and Seungryul Baek.
\newblock {Text2HOI}: Text-guided {3D} motion generation for hand-object interaction.
\newblock In \emph{CVPR}, 2024.

\bibitem[Christen et~al.(2024)Christen, Hampali, Sener, Remelli, Hodan, Sauser, Ma, and Tekin]{christen2024diffh2o}
Sammy Christen, Shreyas Hampali, Fadime Sener, Edoardo Remelli, Tomas Hodan, Eric Sauser, Shugao Ma, and Bugra Tekin.
\newblock {DiffH2O}: Diffusion-based synthesis of hand-object interactions from textual descriptions.
\newblock In \emph{SIGGRAPH Asia}, 2024.

\bibitem[Darcet et~al.(2024)Darcet, Oquab, Mairal, and Bojanowski]{darcet23registers}
Timoth{\'e}e Darcet, Maxime Oquab, Julien Mairal, and Piotr Bojanowski.
\newblock Vision transformers need registers.
\newblock In \emph{ICLR}, 2024.

\bibitem[Delmas et~al.(2022)Delmas, Weinzaepfel, Lucas, Moreno-Noguer, and Rogez]{posescript2022}
Ginger Delmas, Philippe Weinzaepfel, Thomas Lucas, Francesc Moreno-Noguer, and Gr\'{e}gory Rogez.
\newblock {PoseScript}: {3D} human poses from natural language.
\newblock In \emph{ECCV}, 2022.

\bibitem[Dong et~al.(2024)Dong, Chaudhary, Xu, Wang, Lary, and Nwogu]{dong2024signavatar}
Lu Dong, Lipisha Chaudhary, Fei Xu, Xiao Wang, Mason Lary, and Ifeoma Nwogu.
\newblock {SignAvatar}: Sign language {3D} motion reconstruction and generation.
\newblock In \emph{FG}, 2024.

\bibitem[Duarte et~al.(2021)Duarte, Palaskar, Ventura, Ghadiyaram, DeHaan, Metze, Torres, and Gir{\' o}-i Nieto]{duarte2021howsign}
Amanda~Cardoso Duarte, Shruti Palaskar, Lucas Ventura, Deepti Ghadiyaram, Kenneth DeHaan, Florian Metze, Jordi Torres, and Xavier Gir{\' o}-i Nieto.
\newblock {How2Sign}: A large-scale multimodal dataset for continuous {A}merican {S}ign {L}anguage.
\newblock In \emph{CVPR}, 2021.

\bibitem[Dubey et~al.(2024)Dubey, Jauhri, Pandey, Kadian, Al-Dahle, Letman, Mathur, Schelten, Yang, Fan, et~al.]{dubey2024llama}
Abhimanyu Dubey, Abhinav Jauhri, Abhinav Pandey, Abhishek Kadian, Ahmad Al-Dahle, Aiesha Letman, Akhil Mathur, Alan Schelten, Amy Yang, Angela Fan, et~al.
\newblock The {Llama} 3 herd of models.
\newblock \emph{arXiv}, 2024.

\bibitem[Fan et~al.(2023)Fan, Taheri, Tzionas, Kocabas, Kaufmann, Black, and Hilliges]{arctic2023}
Zicong Fan, Omid Taheri, Dimitrios Tzionas, Muhammed Kocabas, Manuel Kaufmann, Michael~J. Black, and Otmar Hilliges.
\newblock {ARCTIC}: A dataset for dexterous bimanual hand-object manipulation.
\newblock In \emph{CVPR}, 2023.

\bibitem[Fenlon et~al.(2014)Fenlon, Cormier, Rentelis, Schembri, Rowley, Adam, and Woll]{signbank2014}
J. Fenlon, K. Cormier, R. Rentelis, A. Schembri, K. Rowley, Robert Adam, and B. Woll.
\newblock {BSL SignBank}: {A} lexical database and dictionary of {British Sign Language} (first edition).
\newblock \emph{London: Deafness, Cognition and Language Research Centre, University College London}, 2014.

\bibitem[Forte et~al.(2023)Forte, Kulits, Huang, Choutas, Tzionas, Kuchenbecker, and Black]{forte2023}
Maria-Paola Forte, Peter Kulits, Chun-Hao~P. Huang, Vasileios Choutas, Dimitrios Tzionas, Katherine~J. Kuchenbecker, and Michael~J. Black.
\newblock Reconstructing signing avatars from video using linguistic priors.
\newblock In \emph{CVPR}, 2023.

\bibitem[{Gemini Team et al.}(2024)]{gemini2024}
{Gemini Team et al.}
\newblock Gemini: A family of highly capable multimodal models.
\newblock \emph{arXiv:2312.11805}, 2024.

\bibitem[Ghosh et~al.(2023)Ghosh, Dabral, Golyanik, Theobalt, and Slusallek]{ghosh2023imos}
Anindita Ghosh, Rishabh Dabral, Vladislav Golyanik, Christian Theobalt, and Philipp Slusallek.
\newblock {IMos}: Intent-driven full-body motion synthesis for human-object interactions.
\newblock In \emph{Eurographics}, 2023.

\bibitem[Ginosar et~al.(2019)Ginosar, Bar, Kohavi, Chan, Owens, and Malik]{ginosar2019gestures}
S. Ginosar, A. Bar, G. Kohavi, C. Chan, A. Owens, and J. Malik.
\newblock Learning individual styles of conversational gesture.
\newblock In \emph{CVPR}, 2019.

\bibitem[Guo et~al.(2022{\natexlab{a}})Guo, Zou, Zuo, Wang, and Ji]{guo2022humanml3d}
Chuan Guo, Shihao Zou, Xinxin Zuo, Sen Wang, and Ji.
\newblock Generating diverse and natural {3D} human motions from text.
\newblock In \emph{CVPR}, 2022{\natexlab{a}}.

\bibitem[Guo et~al.(2022{\natexlab{b}})Guo, Zuo, Wang, and Cheng]{tm2tguo2022}
Chuan Guo, Xinxin Zuo, Sen Wang, and Li Cheng.
\newblock {TM2T}: Stochastic and tokenized modeling for the reciprocal generation of {3D} human motions and texts.
\newblock In \emph{ECCV}, 2022{\natexlab{b}}.

\bibitem[Guo et~al.(2024)Guo, Mu, Javed, Wang, and Cheng]{guo2023momask}
Chuan Guo, Yuxuan Mu, Muhammad~Gohar Javed, Sen Wang, and Li Cheng.
\newblock {MoMask}: Generative masked modeling of {3D} human motions.
\newblock In \emph{CVPR}, 2024.

\bibitem[Ho et~al.(2020)Ho, Jain, and Abbeel]{ho2020denoising}
Jonathan Ho, Ajay Jain, and Pieter Abbeel.
\newblock Denoising diffusion probabilistic models.
\newblock In \emph{NeurIPS}, 2020.

\bibitem[Hwang et~al.(2021)Hwang, Kim, and Park]{hwang2021nonautoregressive}
Eui~Jun Hwang, Jung-Ho Kim, and Jong~C. Park.
\newblock Non-autoregressive sign language production with {G}aussian space.
\newblock In \emph{BMVC}, 2021.

\bibitem[Joo et~al.(2018)Joo, Simon, and Sheikh]{joo2018totalcapture}
Hanbyul Joo, Tomas Simon, and Yaser Sheikh.
\newblock Total capture: A {3D} deformation model for tracking faces, hands, and bodies.
\newblock In \emph{CVPR}, 2018.

\bibitem[Joze and Koller(2019)]{joze2019msasl}
Hamid Reza~Vaezi Joze and Oscar Koller.
\newblock {MS-ASL}: A large-scale data set and benchmark for understanding {A}merican {S}ign {L}anguage.
\newblock In \emph{BMVC}, 2019.

\bibitem[Kingma and Welling(2014)]{kingma2014vae}
Diederik~P Kingma and Max Welling.
\newblock Auto-encoding variational bayes.
\newblock In \emph{ICLR}, 2014.

\bibitem[Kocabas et~al.(2021)Kocabas, Huang, Hilliges, and Black]{kocabas2021pare}
Muhammed Kocabas, Chun-Hao~P. Huang, Otmar Hilliges, and Michael~J. Black.
\newblock {PARE}: Part attention regressor for {3D} human body estimation.
\newblock In \emph{ICCV}, 2021.

\bibitem[Koller et~al.(2015)Koller, Forster, and Ney]{phoenix_cslr_2015}
Oscar Koller, Jens Forster, and Hermann Ney.
\newblock Continuous sign language recognition: Towards large vocabulary statistical recognition systems handling multiple signers.
\newblock \emph{CVIU}, 2015.

\bibitem[Li et~al.(2020)Li, Opazo, Yu, and Li]{li2020wordlevel}
Dongxu Li, Cristian~Rodriguez Opazo, Xin Yu, and Hongdong Li.
\newblock Word-level deep sign language recognition from video: A new large-scale dataset and methods comparison.
\newblock In \emph{WACV}, 2020.

\bibitem[Li et~al.(2023)Li, Tian, Zhang, Wang, Zhang, Bo, Liu, and Chen]{renderIH2023}
Lijun Li, Linrui Tian, Xindi Zhang, Qi Wang, Bang Zhang, Liefeng Bo, Mengyuan Liu, and Chen Chen.
\newblock {RenderIH}: A large-scale synthetic dataset for {3D} interacting hand pose estimation.
\newblock In \emph{ICCV}, 2023.

\bibitem[Lin et~al.(2023{\natexlab{a}})Lin, Zeng, Lu, Cai, Zhang, Wang, and Zhang]{motionx2023}
Jing Lin, Ailing Zeng, Shunlin Lu, Yuanhao Cai, Ruimao Zhang, Haoqian Wang, and Lei Zhang.
\newblock Motion-{X}: A large-scale {3D} expressive whole-body human motion dataset.
\newblock In \emph{NeurIPS}, 2023{\natexlab{a}}.

\bibitem[Lin et~al.(2023{\natexlab{b}})Lin, Zeng, Wang, Zhang, and Li]{lin2023onestage}
Jing Lin, Ailing Zeng, Haoqian Wang, Lei Zhang, and Yu Li.
\newblock One-stage {3D} whole-body mesh recovery with component aware transformer.
\newblock In \emph{CVPR}, 2023{\natexlab{b}}.

\bibitem[Lin et~al.(2023{\natexlab{c}})Lin, Xu, Yang, Liu, Chen, Wang, Yu, and Xu]{lin2023handdiffuse}
Pei Lin, Sihang Xu, Hongdi Yang, Yiran Liu, Xin Chen, Jingya Wang, Jingyi Yu, and Lan Xu.
\newblock {HandDiffuse}: Generative controllers for two-hand interactions via diffusion models.
\newblock \emph{arXiv:2312.04867}, 2023{\natexlab{c}}.

\bibitem[Liu et~al.(2022{\natexlab{a}})Liu, Zhu, Iwamoto, Peng, Li, Zhou, Bozkurt, and Zheng]{beat2022}
Haiyang Liu, Zihao Zhu, Naoya Iwamoto, Yichen Peng, Zhengqing Li, You Zhou, Elif Bozkurt, and Bo Zheng.
\newblock {BEAT}: A large-scale semantic and emotional multi-modal dataset for conversational gestures synthesis.
\newblock In \emph{ECCV}, 2022{\natexlab{a}}.

\bibitem[Liu et~al.(2022{\natexlab{b}})Liu, Ning, Cao, Wei, Zhang, Lin, and Hu]{Liu2022VideoST}
Ze Liu, Jia Ning, Yue Cao, Yixuan Wei, Zheng Zhang, Stephen Lin, and Han Hu.
\newblock Video swin transformer.
\newblock In \emph{CVPR}, 2022{\natexlab{b}}.

\bibitem[Loper et~al.(2015)Loper, Mahmood, Romero, Pons-Moll, and Black]{smpl2015}
Matthew Loper, Naureen Mahmood, Javier Romero, Gerard Pons-Moll, and Michael~J. Black.
\newblock {SMPL}: A skinned multi-person linear model.
\newblock \emph{ACM Transactions on Graphics ({TOG})}, 2015.

\bibitem[Lugaresi et~al.(2019)Lugaresi, Tang, Nash, McClanahan, Uboweja, Hays, Zhang, Chang, Yong, Lee, et~al.]{lugaresi2019mediapipe}
Camillo Lugaresi, Jiuqiang Tang, Hadon Nash, Chris McClanahan, Esha Uboweja, Michael Hays, Fan Zhang, Chuo-Ling Chang, Ming~Guang Yong, Juhyun Lee, et~al.
\newblock Mediapipe: A framework for building perception pipelines.
\newblock \emph{arXiv:1906.08172}, 2019.

\bibitem[Lv et~al.(2024)Lv, Xu, Yan, Jin, Xu, Wu, Liu, Li, Bi, Zeng, and Yang]{lv2024himo}
Xintao Lv, Liang Xu, Yichao Yan, Xin Jin, Congsheng Xu, Shuwen Wu, Yifan Liu, Lincheng Li, Mengxiao Bi, Wenjun Zeng, and Xiaokang Yang.
\newblock {HIMO}: {A} new benchmark for full-body human interacting with multiple objects.
\newblock In \emph{ECCV}, 2024.

\bibitem[Ma et~al.(2024)Ma, Ye, Hong, Guzov, Jiang, Postyeni, Pesqueira, Gamino, Baiyya, Kim, Bailey, Fosas, Liu, Liu, Engel, Nardi, and Newcombe]{ma2024nymeria}
Lingni Ma, Yuting Ye, Fangzhou Hong, Vladimir Guzov, Yifeng Jiang, Rowan Postyeni, Luis Pesqueira, Alexander Gamino, Vijay Baiyya, Hyo~Jin Kim, Kevin Bailey, David~Soriano Fosas, C.~Karen Liu, Ziwei Liu, Jakob Engel, Renzo~De Nardi, and Richard Newcombe.
\newblock Nymeria: A massive collection of multimodal egocentric daily motion in the wild.
\newblock In \emph{ECCV}, 2024.

\bibitem[Mahmood et~al.(2019)Mahmood, Ghorbani, Troje, Pons-Moll, and Black]{amass2019}
Naureen Mahmood, Nima Ghorbani, Nikolaus~F Troje, Gerard Pons-Moll, and Michael~J Black.
\newblock {AMASS: A}rchive of motion capture as surface shapes.
\newblock In \emph{ICCV}, 2019.

\bibitem[Momeni et~al.(2020)Momeni, Varol, Albanie, Afouras, and Zisserman]{momeni20_spotting}
Liliane Momeni, G{\"u}l Varol, Samuel Albanie, Triantafyllos Afouras, and Andrew Zisserman.
\newblock Watch, read and lookup: learning to spot signs from multiple supervisors.
\newblock In \emph{ACCV}, 2020.

\bibitem[Momeni et~al.(2022)Momeni, Bull, Prajwal, Albanie, Varol, and Zisserman]{momeni22bsldensify}
Liliane Momeni, Hannah Bull, K~R Prajwal, Samuel Albanie, G{\"u}l Varol, and Andrew Zisserman.
\newblock Automatic dense annotation of large-vocabulary sign language videos.
\newblock In \emph{ECCV}, 2022.

\bibitem[Moon et~al.(2020)Moon, i~Yu, Wen, Shiratori, and Lee]{interhand26m2020}
Gyeongsik Moon, Shoou i Yu, He Wen, Takaaki Shiratori, and Kyoung~Mu Lee.
\newblock {InterHand2.6M}: A dataset and baseline for {3D} interacting hand pose estimation from a single {RGB} image.
\newblock In \emph{ECCV}, 2020.

\bibitem[Moon et~al.(2022)Moon, Choi, and Lee]{moon2022accurate}
Gyeongsik Moon, Hongsuk Choi, and Kyoung~Mu Lee.
\newblock Accurate {3D} hand pose estimation for whole-body {3D} human mesh estimation.
\newblock In \emph{CVPR}, 2022.

\bibitem[Neidle et~al.(2012)Neidle, Thangali, and Sclaroff]{neidle2012asllvd}
Carol Neidle, Ashwin Thangali, and Stan Sclaroff.
\newblock Challenges in development of the {A}merican {S}ign {L}anguage lexicon video dataset ({ASLLVD}) corpus.
\newblock In \emph{Proc. 5th Workshop on the Representation and Processing of Sign Languages: Interactions between Corpus and Lexicon, LREC}, 2012.

\bibitem[Ng et~al.(2021)Ng, Ginosar, Darrell, and Joo]{body2hands2021}
Evonne Ng, Shiry Ginosar, Trevor Darrell, and Hanbyul Joo.
\newblock {Body2Hands}: Learning to infer {3D} hands from conversational gesture body dynamics.
\newblock In \emph{CVPR}, 2021.

\bibitem[OpenAI(2024)]{gpt4}
OpenAI.
\newblock {GPT}-4 technical report.
\newblock \emph{arXiv:2303.08774}, 2024.

\bibitem[Pavlakos et~al.(2019{\natexlab{a}})Pavlakos, Choutas, Ghorbani, Bolkart, Osman, Tzionas, and Black]{SMPL-X:2019}
Georgios Pavlakos, Vasileios Choutas, Nima Ghorbani, Timo Bolkart, Ahmed A.~A. Osman, Dimitrios Tzionas, and Michael~J. Black.
\newblock Expressive body capture: 3d hands, face, and body from a single image.
\newblock In \emph{Proceedings IEEE Conf. on Computer Vision and Pattern Recognition (CVPR)}, 2019{\natexlab{a}}.

\bibitem[Pavlakos et~al.(2019{\natexlab{b}})Pavlakos, Choutas, Ghorbani, Bolkart, Osman, Tzionas, and Black]{SMPLX2019}
Georgios Pavlakos, Vasileios Choutas, Nima Ghorbani, Timo Bolkart, Ahmed A.~A. Osman, Dimitrios Tzionas, and Michael~J. Black.
\newblock Expressive body capture: {3D} hands, face, and body from a single image.
\newblock In \emph{CVPR}, 2019{\natexlab{b}}.

\bibitem[Pavlakos et~al.(2024)Pavlakos, Shan, Radosavovic, Kanazawa, Fouhey, and Malik]{hamerpavlakos2024}
Georgios Pavlakos, Dandan Shan, Ilija Radosavovic, Angjoo Kanazawa, David Fouhey, and Jitendra Malik.
\newblock Reconstructing hands in {3D} with transformers.
\newblock In \emph{CVPR}, 2024.

\bibitem[Petrovich et~al.(2022)Petrovich, Black, and Varol]{petrovich22temos}
Mathis Petrovich, Michael~J. Black, and G{\"u}l Varol.
\newblock {TEMOS}: Generating diverse human motions from textual descriptions.
\newblock In \emph{ECCV}, 2022.

\bibitem[Petrovich et~al.(2023)Petrovich, Black, and Varol]{petrovich2023tmr}
Mathis Petrovich, Michael~J. Black, and G{\" u}l Varol.
\newblock {TMR}: Text-to-motion retrieval using contrastive {3D} human motion synthesis.
\newblock In \emph{ICCV}, 2023.

\bibitem[Petrovich et~al.(2024)Petrovich, Litany, Iqbal, Black, Varol, Peng, and Rempe]{petrovich24stmc}
Mathis Petrovich, Or Litany, Umar Iqbal, Michael~J. Black, G{\"u}l Varol, Xue~Bin Peng, and Davis Rempe.
\newblock Multi-track timeline control for text-driven {3D} human motion generation.
\newblock In \emph{CVPRW}, 2024.

\bibitem[Plappert et~al.(2016)Plappert, Mandery, and Asfour]{kitml2016}
Matthias Plappert, Christian Mandery, and Tamim Asfour.
\newblock The {KIT} motion-language dataset.
\newblock \emph{Big Data}, 2016.

\bibitem[Potamias et~al.(2025)Potamias, Zhang, Deng, and Zafeiriou]{potamias2025wilor}
Rolandos~Alexandros Potamias, Jinglei Zhang, Jiankang Deng, and Stefanos Zafeiriou.
\newblock {WiLoR}: End-to-end {3D} hand localization and reconstruction in-the-wild.
\newblock In \emph{CVPR}, 2025.

\bibitem[Punnakkal et~al.(2021)Punnakkal, Chandrasekaran, Athanasiou, Quiros-Ramirez, and Black]{babel2021}
Abhinanda~R. Punnakkal, Arjun Chandrasekaran, Nikos Athanasiou, Alejandra Quiros-Ramirez, and Michael~J. Black.
\newblock {BABEL}: Bodies, action and behavior with english labels.
\newblock In \emph{CVPR}, 2021.

\bibitem[Qi et~al.(2023)Qi, Liu, Sun, Li, Fan, and Yu]{qi2023}
Xingqun Qi, Chen Liu, Muyi Sun, Lincheng Li, Changjie Fan, and Xin Yu.
\newblock Diverse {3D} hand gesture prediction from body dynamics by bilateral hand disentanglement.
\newblock In \emph{CVPR}, 2023.

\bibitem[Qi et~al.(2024)Qi, Liu, Li, Hou, Xin, and Yu]{emotiongesture2024}
Xingqun Qi, Chen Liu, Lincheng Li, Jie Hou, Haoran Xin, and Xin Yu.
\newblock {EmotionGesture}: Audio-driven diverse emotional co-speech {3D} gesture generation.
\newblock \emph{IEEE Transactions on Multimedia}, 2024.

\bibitem[Radford et~al.(2021)Radford, Kim, Hallacy, Ramesh, Goh, Agarwal, Sastry, Askell, Mishkin, Clark, et~al.]{radford2021clip}
Alec Radford, Jong~Wook Kim, Chris Hallacy, Aditya Ramesh, Gabriel Goh, Sandhini Agarwal, Girish Sastry, Amanda Askell, Pamela Mishkin, Jack Clark, et~al.
\newblock Learning transferable visual models from natural language supervision.
\newblock In \emph{ICML}, 2021.

\bibitem[Raude et~al.(2024)Raude, Prajwal, Momeni, Bull, Albanie, Zisserman, and Varol]{raude24cslr2}
Charles Raude, K.~R. Prajwal, Liliane Momeni, Hannah Bull, Samuel Albanie, Andrew Zisserman, and G{\" u}l Varol.
\newblock A tale of two languages: Large-vocabulary continuous sign language recognition from spoken language supervision.
\newblock \emph{arXiv}, 2024.

\bibitem[Romero et~al.(2017)Romero, Tzionas, and Black]{MANO2017}
Javier Romero, Dimitrios Tzionas, and Michael~J. Black.
\newblock Embodied hands: Modeling and capturing hands and bodies together.
\newblock In \emph{SIGGRAPH Asia}, 2017.

\bibitem[Sanh et~al.(2019)Sanh, Debut, Chaumond, and Wolf]{sanh2019distilbert}
Victor Sanh, Lysandre Debut, Julien Chaumond, and Thomas Wolf.
\newblock {DistilBERT}, a distilled version of {BERT}: smaller, faster, cheaper and lighter.
\newblock In \emph{Energy Efficient Machine Learning and Cognitive Computing {NeurIPS} Workshop}, 2019.

\bibitem[Saunders et~al.(2020{\natexlab{a}})Saunders, Bowden, and Camg{\" o}z]{saunders2020adversarial}
Ben Saunders, Richard Bowden, and Necati~Cihan Camg{\" o}z.
\newblock Adversarial training for multi-channel sign language production.
\newblock In \emph{BMVC}, 2020{\natexlab{a}}.

\bibitem[Saunders et~al.(2020{\natexlab{b}})Saunders, Camg{\" o}z, and Bowden]{saunders2020progressive}
Ben Saunders, Necati~Cihan Camg{\" o}z, and Richard Bowden.
\newblock Progressive transformers for end-to-end sign language production.
\newblock In \emph{ECCV}, 2020{\natexlab{b}}.

\bibitem[Saunders et~al.(2021)Saunders, Camg{\" o}z, and Bowden]{saunders2021mixed}
Ben Saunders, Necati~Cihan Camg{\" o}z, and Richard Bowden.
\newblock Mixed {SIGNals}: Sign language production via a mixture of motion primitives.
\newblock In \emph{ICCV}, 2021.

\bibitem[Saunders et~al.(2022)Saunders, Camg{\" o}z, and Bowden]{saunders2022signing}
Ben Saunders, Necati~Cihan Camg{\" o}z, and Richard Bowden.
\newblock Signing at scale: Learning to co-articulate signs for large-scale photo-realistic sign language production.
\newblock In \emph{CVPR}, 2022.

\bibitem[Stoll et~al.(2018)Stoll, Camg{\" o}z, Hadfield, and Bowden]{stoll2018sign}
Stephanie Stoll, Necati~Cihan Camg{\" o}z, Simon Hadfield, and Richard Bowden.
\newblock Sign language production using neural machine translation and generative adversarial networks.
\newblock In \emph{BMVC}, 2018.

\bibitem[Stoll et~al.(2020)Stoll, Camgoz, Hadfield, and Bowden]{stoll2020textsign}
Stephanie Stoll, Necati~Cihan Camgoz, Simon Hadfield, and Richard Bowden.
\newblock {Text2Sign}: Towards sign language production using neural machine translation and generative adversarial networks.
\newblock \emph{IJCV}, 2020.

\bibitem[Stoll et~al.(2022)Stoll, Mustafa, and Guillemaut]{stoll2022there}
Stephanie Stoll, Armin Mustafa, and Jean-Yves Guillemaut.
\newblock There and back again: {3D} sign language generation from text using back-translation.
\newblock In \emph{3DV}, 2022.

\bibitem[Taheri et~al.(2020)Taheri, Ghorbani, Black, and Tzionas]{GRAB2020}
Omid Taheri, Nima Ghorbani, Michael~J. Black, and Dimitrios Tzionas.
\newblock {GRAB}: A dataset of whole-body human grasping of objects.
\newblock In \emph{ECCV}, 2020.

\bibitem[Taheri et~al.(2022)Taheri, Choutas, Black, and Tzionas]{taheri2022goal}
Omid Taheri, Vasileios Choutas, Michael~J. Black, and Dimitrios Tzionas.
\newblock {GOAL}: Generating {4D} whole-body motion for hand-object grasping.
\newblock In \emph{CVPR}, 2022.

\bibitem[Taheri et~al.(2024)Taheri, Zhou, Tzionas, Zhou, Ceylan, Pirk, and Black]{taheri2024grip}
Omid Taheri, Yi Zhou, Dimitrios Tzionas, Yang Zhou, Duygu Ceylan, Soren Pirk, and Michael~J. Black.
\newblock {GRIP}: Generating interaction poses using spatial cues and latent consistency.
\newblock In \emph{3DV}, 2024.

\bibitem[Tevet et~al.(2022)Tevet, Gordon, Hertz, Bermano, and Cohen-Or]{tevet2022motionclip}
Guy Tevet, Brian Gordon, Amir Hertz, Amit~H Bermano, and Daniel Cohen-Or.
\newblock {MotionCLIP}: Exposing human motion generation to {CLIP} space.
\newblock In \emph{ECCV}, 2022.

\bibitem[Tevet et~al.(2023)Tevet, Raab, Gordon, Shafir, Cohen-Or, and Bermano]{mdmtevet2023}
Guy Tevet, Sigal Raab, Brian Gordon, Yonatan Shafir, Daniel Cohen-Or, and Amit~Haim Bermano.
\newblock Human motion diffusion model.
\newblock In \emph{ICLR}, 2023.

\bibitem[Varol et~al.(2021)Varol, Momeni, Albanie, Afouras, and Zisserman]{varol21_bslattend}
G{\"u}l Varol, Liliane Momeni, Samuel Albanie, Triantafyllos Afouras, and Andrew Zisserman.
\newblock Read and attend: Temporal localisation in sign language videos.
\newblock In \emph{CVPR}, 2021.

\bibitem[Ventura et~al.(2020)Ventura, Duarte, and Gir{\' o}-i Nieto]{ventura2020everybody}
Lucas Ventura, Amanda~Cardoso Duarte, and Xavier Gir{\' o}-i Nieto.
\newblock Can everybody sign now? {E}xploring sign language video generation from {2D} poses.
\newblock In \emph{ECCVW}, 2020.

\bibitem[Xu et~al.(2020)Xu, Bazavan, Zanfir, Freeman, Sukthankar, and Sminchisescu]{ghum2020}
Hongyi Xu, Eduard~Gabriel Bazavan, Andrei Zanfir, Bill Freeman, Rahul Sukthankar, and Cristian Sminchisescu.
\newblock {GHUM} \& {GHUML}: Generative {3D} human shape and articulated pose models.
\newblock In \emph{CVPR}, 2020.

\bibitem[Xu et~al.(2024)Xu, Lv, Yan, Jin, Wu, Xu, Liu, Zhou, Rao, Sheng, Liu, Zeng, and Yang]{xu2023inter}
Liang Xu, Xintao Lv, Yichao Yan, Xin Jin, Shuwen Wu, Congsheng Xu, Yifan Liu, Yizhou Zhou, Fengyun Rao, Xingdong Sheng, Yunhui Liu, Wenjun Zeng, and Xiaokang Yang.
\newblock Inter-x: Towards versatile human-human interaction analysis.
\newblock In \emph{CVPR}, 2024.

\bibitem[Yin et~al.(2023)Yin, Wang, He, Liu, Zhao, Li, Jin, and Lin]{emog2023}
Lianying Yin, Yijun Wang, Tianyu He, Jinming Liu, Wei Zhao, Bohan Li, Xin Jin, and Jianxin Lin.
\newblock {EMoG}: Synthesizing emotive co-speech {3D} gesture with diffusion model.
\newblock \emph{arXiv}, 2023.

\bibitem[Yu et~al.(2023)Yu, Huang, Fang, Breckon, and Wang]{yu2023acr}
Zhengdi Yu, Shaoli Huang, Chen Fang, Toby~P. Breckon, and Jue Wang.
\newblock {ACR}: Attention collaboration-based regressor for arbitrary two-hand reconstruction.
\newblock In \emph{CVPR}, 2023.

\bibitem[Yu et~al.(2024)Yu, Huang, Cheng, and Birdal]{signavatars2024}
Zhengdi Yu, Shaoli Huang, Yongkang Cheng, and Tolga Birdal.
\newblock {SignAvatars}: A large-scale {3D} sign language holistic motion dataset and benchmark.
\newblock In \emph{ECCV}, 2024.

\bibitem[Zelinka and Kanis(2020)]{zelinka2020neural}
Jan Zelinka and Jakub Kanis.
\newblock Neural sign language synthesis: Words are our glosses.
\newblock In \emph{WACV}, 2020.

\bibitem[Zhang et~al.(2023)Zhang, Cai, Pan, and Hong]{motiondiffusezhang2022}
Mingyuan Zhang, Zhongang Cai, Liang Pan, and Hong.
\newblock {MotionDiffuse}: Text-driven human motion generation with diffusion model.
\newblock \emph{TPAMI}, 2023.

\bibitem[Zhang et~al.(2024)Zhang, Huang, Zhou, Zhang, Yu, Wang, and Xu]{both2hands2024}
Wenqian Zhang, Molin Huang, Yuxuan Zhou, Juze Zhang, Jingyi Yu, Jingya Wang, and Lan Xu.
\newblock {BOTH2Hands}: Inferring {3D} hands from both text prompts and body dynamics.
\newblock In \emph{CVPR}, 2024.

\bibitem[Zhou et~al.(2021)Zhou, Zhou, Qi, Pu, and Li]{zhou2021csldaily}
Hao Zhou, Wengang Zhou, Weizhen Qi, Junfu Pu, and Houqiang Li.
\newblock Improving sign language translation with monolingual data by sign back-translation.
\newblock In \emph{CVPR}, 2021.

\bibitem[Zhou et~al.(2022)Zhou, Bhatnagar, Lenssen, and Pons-Moll]{zhou2023toch}
Keyang Zhou, Bharat~Lal Bhatnagar, Jan~Eric Lenssen, and Gerard Pons-Moll.
\newblock {TOCH}: Spatio-temporal object-to-hand correspondence for motion refinement.
\newblock In \emph{ECCV}, 2022.

\bibitem[Zhu et~al.(2023)Zhu, Liu, Liu, Qian, Liu, and Yu]{zhu2023}
Lingting Zhu, Xian Liu, Xuanyu Liu, Rui Qian, Ziwei Liu, and Lequan Yu.
\newblock Taming diffusion models for audio-driven co-speech gesture generation.
\newblock In \emph{CVPR}, 2023.

\bibitem[Zuo et~al.(2024)Zuo, Potamias, Ververas, Deng, and Zafeiriou]{zuo2024signstokens}
Ronglai Zuo, Rolandos~Alexandros Potamias, Evangelos Ververas, Jiankang Deng, and Stefanos Zafeiriou.
\newblock Signs as tokens: An autoregressive multilingual sign language generator.
\newblock \emph{arXiv:2411.17799}, 2024.

\end{thebibliography}
}

\clearpage
{\noindent \large \bf {APPENDIX}}\\
\appendix

\renewcommand{\thefigure}{A.\arabic{figure}} %
\setcounter{figure}{0} 
\renewcommand{\thetable}{A.\arabic{table}}
\setcounter{table}{0}

This appendix provides
a brief glossary of terminology (\cref{subsec:app:glossary}),
an analysis of our \dataset dataset (\cref{sec:app:analysis}),
implementation details (\cref{sec:app:details}),
additional qualitative and quantitative results (\cref{sec:app:experiments}).
\appendixref{We also attach a supplementary video to visualize dynamic motions
(see \texttt{video.mp4})}
{\changed{We also provide a video on our project page to visualize dynamic motions}}.

\changed{
	\section{Glossary}
	\label{subsec:app:glossary}
	We list a brief glossary to describe several sign language specific terminology used in the main paper:\\	
	$\bullet$~\textbf{Gloss:} The written representation of a %
	sign, typically a single word. In linguistic glossing, the transcription follows certain rules (e.g., adding `PT:' prefix for the pointing sign `YOU', assigning a unique gloss ID to each sign variant); however, in this paper, we simplify as in \cite{raude24cslr2} and abuse the 
	gloss terminology: the sign-level annotations we use are not careful linguistic glosses, but rather free-form sign-level translations. \\
	$\bullet$~\textbf{Phonology tags:}
	The set of handshapes, movements, locations used to construct the signs.
	See \appendixref{Sec.~3.1}{\cref{subsec:data}} of the main paper and {\cref{subsec:app:signbank}} for the specific list of attributes included in our study.\\
	$\bullet$~\textbf{Sign variant:}
	One of the different ways of signing the same word.
	There may be multiple ways to sign a word due to several reasons.
	We give explanations and examples in \cref{subsec:app:assignment}.
}

\changed{
\section{\dataset Dataset Analysis}
\label{sec:app:analysis}

We analyze our large-scale \dataset dataset
by reporting several statistics (\cref{subsec:app:datastats}),
and estimating the noise level (\cref{subsec:app:noise}).
We also provide more details about our manually cleaned test set verification
(\cref{subsec:app:testset}).

\subsection{Statistics} %
\label{subsec:app:datastats}

In \cref{fig:app:wordspertext}, we plot the distribution of
word count
per unique text descriptions. We show two separate histograms for %
(i)~the texts generated from SignBank phonology attributes and
(ii)~the ones generated from SignBank attributes combined with HMS.
The difference between the mean values of the two distributions remains small ($\mu_{\text{[Phonology]}}=40.2$ and $\mu_{\text{[Phonology+HMS]}}=41.0$), despite the latter containing more information.
Phonology-only descriptions close the gap by containing generic movement words such as `the hand moves'.

In \cref{fig:app:samplespertext}, we also plot the number of motion samples per unique text description in our LLM(Phonology+HMS) training set.
Given that the glosses have a long-tailed distribution in BOBSL, the motions are not uniformly distributed across descriptions either.
The dataset includes 17,552 unique descriptions ($x$-axis), among which 8,201 come from %
Phonology-only
and 9,351 come from
the combination with HMS.

\begin{figure}
	\centering
	\includegraphics[width=.9\linewidth]{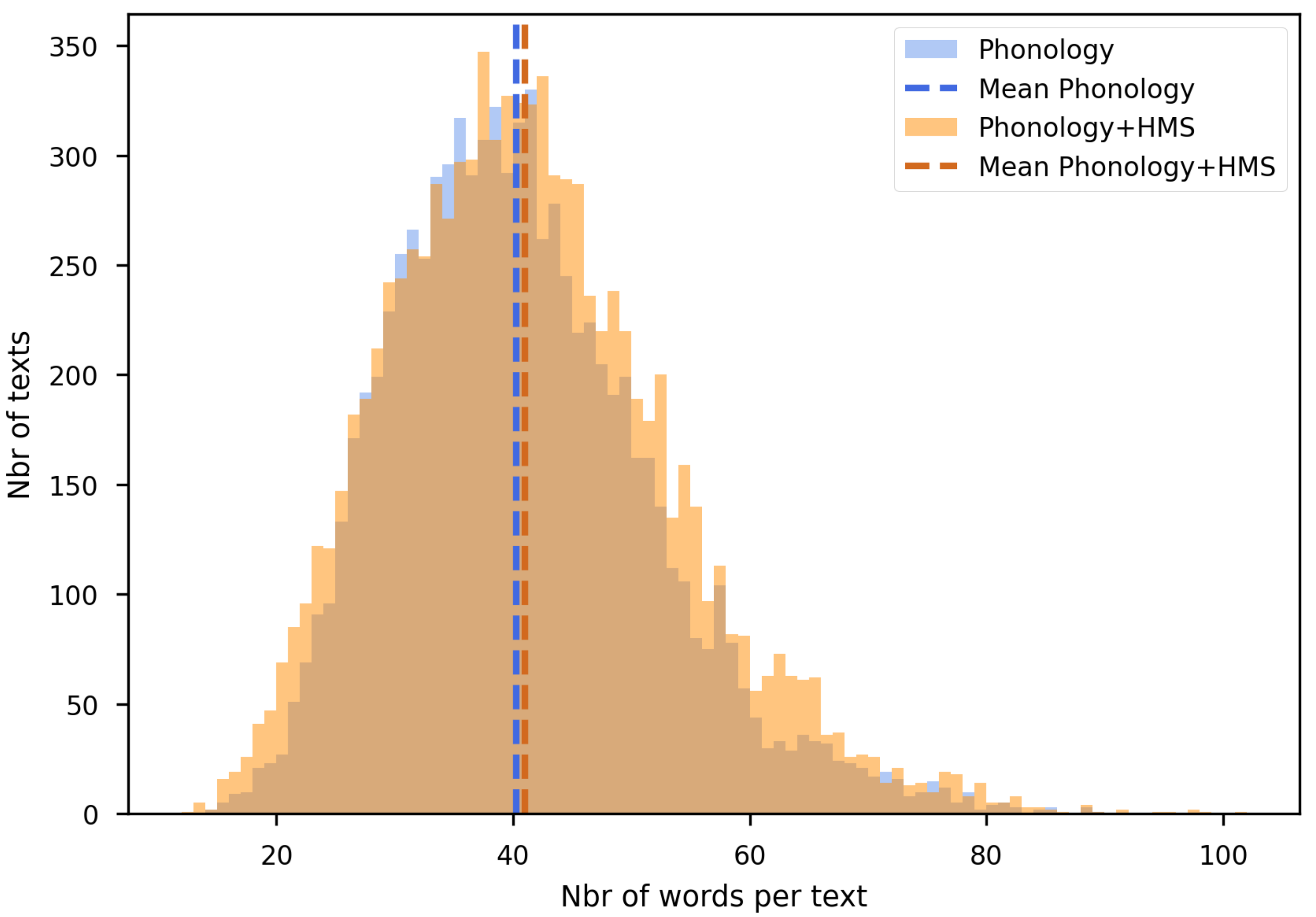}
	\vspace{-0.3cm}
	\caption{\textbf{Histograms of the number of words per text in \dataset training set:} We report the distribution of words per unique text, for the descriptions generated from SignBank phonology, and the descriptions generated from SignBank phonology combined with HMS separately. %
	}
	\label{fig:app:wordspertext}
\end{figure}

\begin{figure}
	\centering
	\includegraphics[width=.9\linewidth]{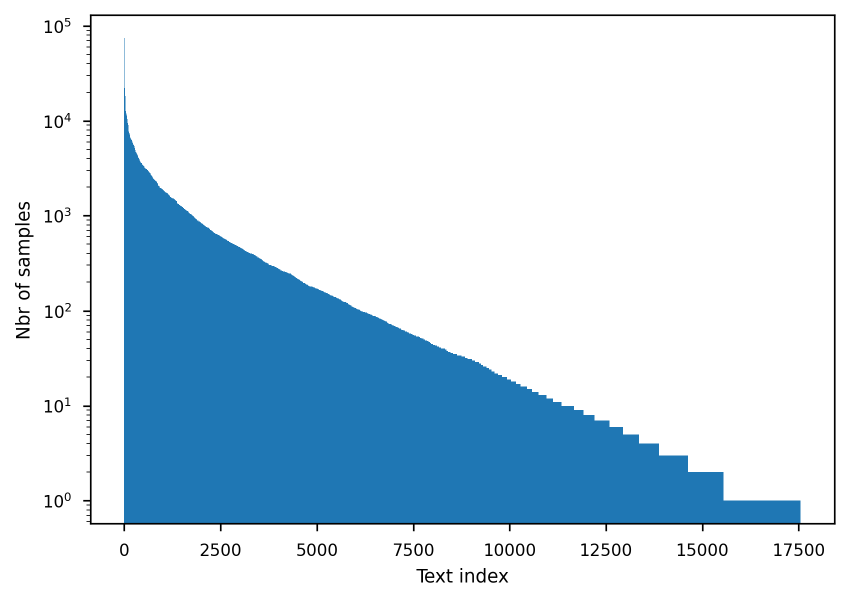}
	\vspace{-0.3cm}
	\caption{\textbf{Number of motion samples per text description:} We observe a long-tailed distribution on the \dataset training set. The median is 33 motion samples per text.}
	\label{fig:app:samplespertext}
\end{figure}

We note two details to help interpreting these numbers.
(a)~Original LLM outputs from SignBank are generated with the dominant/non-dominant hand terminology,
which are then replaced with the words right/left when assigning to BOBSL signs depending on the metadata on right-/left-handedness of signers (i.e., a left handed person's dominant hand would be the left hand with mirrored signing).
With the original dominant terminology, we have 6,090 (Phonology) and 7,032 (Phonology+HMS) unique text descriptions on SignBank,
which then increase to the above numbers after replacing with left/right words during BOBSL assignment.
(b)~The source of these descriptions are 2,351 glosses. Each gloss is converted to 3 different LLM descriptions (2,351 $\times$ 3 = 7,053). Due to some glosses sharing the exact same set of attributes, we have slightly lower number of unique descriptions than 7,053 (e.g., 6,090 for Phonology portion of the dataset).

\subsection{Noise estimation} %
\label{subsec:app:noise}
We quantify the noise on a training subset of 2,525
samples by
manually checking whether the motion corresponds to the text.
We find that
61\% of the motion samples %
correspond to their descriptions.
The main sources of noise are pseudo-glosses
(e.g., 60\% of 2,624 manually-glossed test set match the pseudo-glosses),
and potentially the SignBank variant assignment.
We also note that the 3D pose estimates contribute to the level of noise,
by not strictly following the video, but our analysis here focuses on the fidelity to the text
rather than the motion realism.

\subsection{Test set verification} %
\label{subsec:app:testset}

As mentioned in \appendixref{Sec.~4.1}{\cref{subsec:eval}} of the main paper,
we manually verify the entire test set (seen and unseen) as follows.
We first automatically assign a SignBank variant to each BOBSL test motion, by selecting the variant with the highest cosine similarity in the THMR embedding space.
Then, for each of these SignBank samples,
we collect %
4 most similar BOBSL motions,
and manually select one of them
that matches the SignBank motion, or discard the gloss if we find no match.

We also manually modified the LLM(Phonolgy+HMS) output descriptions for a subset of the test samples,
to make sure the text corresponds to the motion. Due to its smaller size, we have done this correction only on the unseen test set.

Note that the entire test set was already manually \textit{glossed} (as opposed to the training set);
therefore, we did not need to verify the correctness of
matching between the videos and the glosses.
}
\section{Implementation Details}
\label{sec:app:details}
We give additional information about
\methodname training (\cref{subsec:app:hmdm}),
THMR training (\cref{subsec:app:thmr}),
stitching of HAMER hands to SMPLer-X body (\cref{subsec:app:stitching}),
SignBank dictionary attributes (\cref{subsec:app:signbank}),
HandMotionScript (\cref{subsec:app:hms}),
the LLM prompt (\cref{subsec:app:prompt}),
and 
the assignment between SignBank and BOBSL (\cref{subsec:app:assignment}).

\changed{
\subsection{\methodname training and inference} %
\label{subsec:app:hmdm}
Similar to~\cite{petrovich24stmc},
we train the \methodname networks
using a learning rate of ${10^{-4}}$ with AdamW optimizer and a batch size of 32.
With BOBSL3DT, we train \methodname for 300 epochs and use a guidance coefficient of $\lambda = 15$ at test time. With ASL-Text and MS-ZSSLR, we train for 2000 epochs and test with $\lambda = 5$. With BOTH57M, we train for 300 epochs and %
test with $\lambda = 15$. %
}

\subsection{THMR training} %
\label{subsec:app:thmr}
Similar to TMR for bodies~\cite{petrovich2023tmr},
we train the THMR networks for hands
using a learning rate of ${10^{-4}}$ with AdamW optimizer and a batch size of 32.
DistilBERT~\cite{sanh2019distilbert} text embeddings are used as input to the text encoder.
The latent dimensionality of the embedding space is 256, the temperature $\tau=0.1$, and the contrastive loss weight $\lambda_{NCE}=0.1$. The network is trained for 100 epochs.
\changed{As THMR is used for evaluation purposes,
to obtain the best possible performance,	
we use for supervision the 1.3M motion-text samples from our \dataset training data \textit{prior} to filtering the unseen signs.} %
The input format is the same as \methodname 
(see \appendixref{Sec.~3.2}{\cref{subsec:training}}).

\noindent\changed{\textbf{Performance of THMR on real motions.} We report THMR performances on ground truth motions in the first row of \cref{tab:app:input_m2t}. Note that the unseen set is not unseen with regards to THMR
as mentioned above. We refer to \cref{subsec:app:m2t} for comments on the results.}

\noindent\textbf{THMR for transfer experiments.}
\changed{For ASL-Text and MS-ZSSLR retrieval score evaluation, we train THMR only on the arm and hand features,
discarding the torso.
We notice that a significant portion of these datasets include seated signers, where the torso position creates a domain gap with \dataset containing standing signers. Additionally the 3D reconstruction for seated signers tend to estimate unrealistic torso bending for the ground truth, which also motivated our exclusion of the torso features.
Specifically we use the following 6D joint rotations: 2 shoulders, 2 elbows, 2 wrists and 30 for hands, resulting in a total of 216 input features.

BOTH57M full body motions are a lot more dynamic than in sign language datasets, and the textual descriptions for hands only describe finger movements (not arms). Therefore, for the evaluation on BOTH57M, we further reduce the training features of THMR to the 30 hand 6D rotations, resulting in a total of 180 features.}

\begin{figure*}
	\centering
	\includegraphics[width=.9\linewidth]{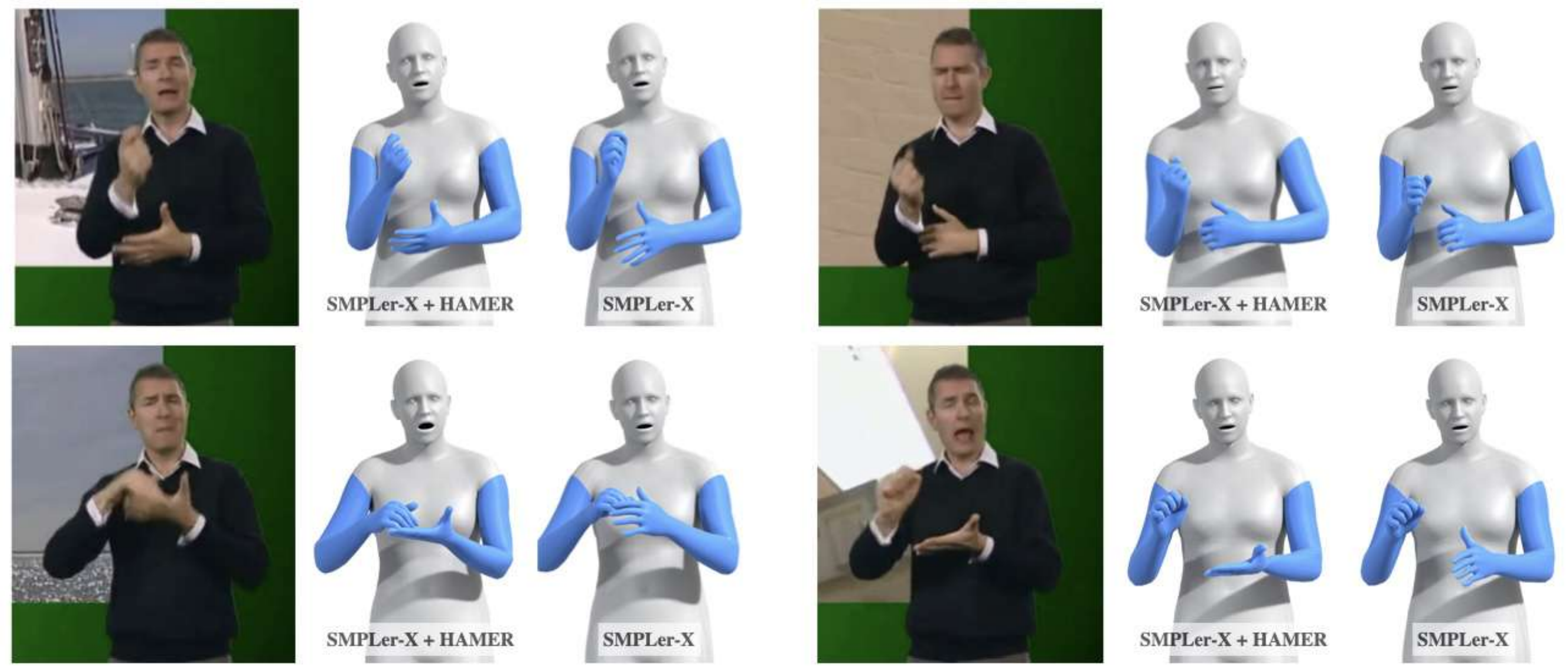}
	\vspace{-0.3cm}
	\caption{\textbf{Stitching HAMER hands to SMPLer-X body:} We observe qualitatively better performance
		of capturing hands when using HAMER~\cite{hamerpavlakos2024}, as opposed to the hand components of SMPLer-X~\cite{cai2023smplerx}. However, HAMER only provides flying hands, which we stitch into
		SMPL-X bodies estimated by SMPLer-X.
	}
	\vspace{-0.3cm}
	\label{fig:app:stitching}
\end{figure*}

\subsection{Stitching HAMER hands to SMPLer-X body}
\label{subsec:app:stitching}
As explained in 
\appendixref{Sec.~3.1}{\cref{subsec:data}} of the main paper,
we use a combination of HAMER~\cite{hamerpavlakos2024}
and SMPLer-X~\cite{cai2023smplerx} for estimating 3D bodies
with hands from videos.
In \cref{fig:app:stitching}, we provide SMPLer-X results
on BOBSL videos with and without using HAMER,
and observe consistently more reliable estimates with
the hand-specific and more recent approach of HAMER.
However, in terms of the global location of wrists, we find
SMPLer-X to be better since it explicitly estimates the rest of the body,
and arm movements define the hand location.

In more detail, HAMER gives us 30 hand joint rotations (local rotations
with respect to the parent in the kinematic tree), which we use
to replace those estimated by SMPLer-X.
For the wrists, we use the global wrist orientation estimated from 
HAMER which are in world coordinate with the origin at the wrist.
We convert these to be local wrist orientations and replace those
estimated by SMPLer-X.
\changed{We first stitch both estimates at the wrists by using \mbox{SMPLer-X} wrist positions and HAMER wrist global orientations. This provides realistic joint positions but can lead to unrealistic wrist rotations. For instance, if SMPLer-X estimates a hand facing down but HAMER estimates a hand facing up, this first stitching results in a local rotation at the wrist of about 180 degrees, which is unrealistic. 

Therefore, inspired by~\cite{SMPL-X:2019} and~\cite{xu2023inter}, we then perform optimization to adjust the SMPL-X rotations of the arm joints.  
We optimize the shoulders, elbows and wrists (6 rotations)
by minimizing the Euclidean distance between the 3D joints derived from the new rotations and the initial stitched pose.
Specifically, the target coordinates are the 3D coordinates of the elbows (2 joints), wrists (2 joints) and hands (30 joints). We initialize the target joint rotations with the SMPL neutral pose for each frame, i.e., with the arms positioned in a T-pose. Similar to~\cite{xu2023inter}, we add a temporal smoothing term that minimizes the Euclidean distance between the joint positions from consecutive frames. A regularization term also minimizes the magnitude of the rotations to avoid unrealistic bending.} %

\subsection{SignBank dictionary attributes}
\label{subsec:app:signbank}
We refer to \cref{fig:app:sb1} and \cref{fig:app:sb2}
to illustrate complete examples for the dictionary attributes.
Beyond those mentioned in
\appendixref{Sec.~3.1}{\cref{subsec:data}} of the main paper, 
phonological information also contains whether signs are performed with only one hand, or with both hands.
In case of two-handed signs, the phonology also denotes 
whether the sign is `symmetric' or `alternating',
and whether one of the hands is static.
There may also be information on whether forearm rotations are involved,
and whether a handshape change occurs during the motion 
(in addition to the initial/final handshapes).

\changed{To transform these SignBank attributes into our LLM input, we replace linguistic terms in field names and categories with more descriptive English descriptions. Additionally, %
we provide several descriptions for an attribute
where possible,
as
text augmentation. We randomly select from these attribute descriptions for each gloss when inputting the attributes to the LLM. For instance, for the handshape attributes, `flat' will be converted into a description among [\textit{`Flat', `The hand is held flat with the fingers held together.', `The fingers are extended and together.'}], and `small' will be converted into a description among [\textit{`The index finger is extended and bent inward. The thumb is extended, parallel to the index finger. The other fingers are curled on the palm.', `The extended index finger, bent at the palm knuckle, and extended thumb are held parallel to each other.'}]; for the tags, `double handed' will be converted into the description \textit{`both hands are used, with the same handshape on both hands'} and `two handed' into \textit{`both hands are used, and their motions aren't symmetrical'}.  Specifically,
we take part of the handshape descriptions
from~\cite{brien1992bsldict}.
The full mapping can be found in \appendixref{\texttt{mapping\_values.py} in our supplementary files}{our code release}.
}

\subsection{HandMotionScript}
\label{subsec:app:hms}
We provide
a pseudo-code in \cref{alg:app:hms}
for our HandMotionScript (HMS) pipeline,
and qualitative examples in 
\cref{fig:app:sb1} and \cref{fig:app:sb2}.
We build on PoseScript~\cite{posescript2022}
designed for full human bodies.	
From PoseScript features, we use the hand \textbf{distances} to the relevant body parts,
determined by SignBank attributes.
If the hand location attribute is not the `neutral space', but a specific body part, we compute the distances
from the hands to this body part, to detect the movement features.
If the motion is two-handed, we compute distances between the two hands. If the motion is one-handed, we compute the distances for the signer's dominant hand only.
We further break down the distances into the \textbf{3 axes (x-y-z)}
to provide more detailed information, such as left/right or above/below relationships.
As in PoseScript, we convert numerical values into text codes by assigning a range of values for each feature.
We adapt some of these ranges for our HMS pipeline.
\changed{We use the following text codes for the 4 distance features:\\
$\bullet$~\textbf{Distance:} [\textit{touching, close, medium, spread, wide}]\\
$\bullet$~\textbf{x-coordinate:} [\textit{wide/right, spread/right, medium/right, close/right, touching, close/left, medium/left, spread/left, wide/left}].\\
$\bullet$~\textbf{y-coordinate:} [\textit{wide/below, spread/below, medium/below, close/below, touching, close/above, medium/above, spread/above, wide/above}].\\
$\bullet$~\textbf{z-coordinate:} [\textit{wide/behind, spread/behind, medium/behind, close/behind, touching, close/in front, medium/in front, spread/in front, wide/in front}].
}
    
We also add a new feature to PoseScript to extract the \textbf{palm orientations}.
Specifically, \changed{for each hand, we compute the vector normal to the wrist global orientation. If one of its global coordinate norms is greater than a chosen threshold (0.7), we assign the axis of this coordinate as the palm direction. The sign of the coordinate determines the final orientation. A palm can thus be assigned an orientation among the following possibilities, respectively to the 3 axes: %
\textit{sideways, down/up, in/out}.}

\changed{Once the above frame-level features are extracted, we apply the following post-processing in order to shorten them and remove possible noise. We remove every text code with less than 4 consecutive repetitions, and
we collapse consecutive repetitions into a single text code.
Furthermore, when a distance feature has multiple text codes for multiple axes, we discard the axis-dependent distance features (x/y/z) entirely to prevent too lengthy and noisy descriptions from the LLM.
Finally,
PoseScript wording uses `right' and `left' hands, whereas our raw SignBank descriptions follow `dominant' and `non-dominant' terminology (see \cref{subsec:app:datastats}).
We replace the right/left words with dominant/non-dominant according to the SignBank tags which give information on whether a signer is left-handed or right-handed.
The temporal sequence of words are then given to the LLM. }

\begin{algorithm}
\caption{HandMotionScript pipeline}
\begin{algorithmic}[1]
    \State \textbf{Input:} 3D motion world coordinates (n frames)
    \State \textbf{Input:} Global hand orientation (n frames)
    \State \textbf{Input:} SignBank location attributes (SBlocs)
    \For{frame $\gets 1$ to $n$}
        \For{SBlocs $\gets$ loc}
            \If{loc is Neutral Space}
                loc $\gets$ other hand
            \EndIf
            \If{left hand is used}
                \State Compute distance posecodes: left hand to loc
            \EndIf
            \If{right hand is used}
                \State Compute distance posecodes: right hand to loc
            \EndIf
        \EndFor
        \If{left hand is used}
            \State Compute orientation posecodes: left hand
        \EndIf
        \If{right hand is used}
            \State Compute orientation posecodes: right hand
        \EndIf
    \EndFor
    \State \textbf{Return} Distance posecode sequences (n frames)
    \State \textbf{Return} Orientation posecode sequences (n frames)
\end{algorithmic}
\label{alg:app:hms}
\end{algorithm}

\subsection{LLM prompting}
\label{subsec:app:prompt}
We use Gemini 2.5 Pro \textit{\mbox{Thinking Experimental 01-21}}~\cite{gemini2024}
as our LLM.
\changed{We provide
the exact prompt for generating the textual descriptions from the SignBank and HMS attributes in 
\cref{alg:app:prompt1,alg:app:prompt2,alg:app:prompt3,alg:app:prompt4,alg:app:prompt5}.
The prompt}
contains the task instructions and few-shot (\changed{six}) examples for in-context learning.
We also include the prompt version when HMS is not used in \changed{\appendixref{our supplementary files
(\texttt{prompt\_phonology.txt} and \texttt{prompt\_phonology\_HMS.txt})}{our project page}.}
We use both prompts to construct our \dataset dataset.

\noindent\changed{\textbf{LLM output post-processing.}
For single-handed signs, we %
remove any mention of the non-dominant hand `remaining still' from the LLM output.
This is because in BOBSL, the non-dominant hand is rarely still (due to the natural signing speed) %
as opposed to the slow SignBank videos.

Furthermore, once the SignBank LLM-generated texts are assigned to BOBSL motions, we replace the words `dominant' and `non-dominant' %
with `left' or `right' depending on whether the signer is left-handed or right-handed using the metadata.}

\subsection{SignBank gloss assignment for BOBSL motions}
\label{subsec:app:assignment}
Each BOBSL video corresponds to one pseudo-gloss, represented as a word (or more rarely a phrase).
On the other hand, SignBank is a linguistic dictionary, where each gloss corresponds to a unique
sign variant, characterized by one motion.
For example, the word `happy' corresponds to three glosses HAPPY, HAPPYb, HAPPYc.
The mapping between words and glosses is complex~\cite{raude24cslr2}.
A word can map to multiple SignBank glosses,
and a gloss can map to multiple words.
These may be due to
multiple ways of signing the same concept,
due to homonyms (same sign motion with different meanings,
such as BSL signs for `battery' and `uncle'),
or due to English synoyms (`happy' and `content').

SignBank also includes \textit{keywords} for each gloss,
related concepts which may share the same signing.
For example, the keywords for the gloss `HAPPY'
include `happiness', `merry', `enjoy', `fun', while
they include `excited', `jump for joy' for the gloss `HAPPYc'.
When constructing the list of \textit{variants} for a given BOBSL pseudo-gloss word (e.g., `happy'),
we consider all possible keywords (e.g., `fun', therefore FUN, FUNb, FUNc...),
as well as all glosses that map to this word (e.g., `HAPPYc').

Once we construct a list of candidate gloss motions,
\changed{for each
	sign
	label in BOBSL, we perform k-medoids clustering on the set of its paired BOBSL motions combined with each candidate SignBank variant motion. We initialize the number of clusters as the number of available SignBank variants. For each cluster thus obtained, if the cluster includes: (i) a unique SignBank variant, it is assigned to each of its BOBSL samples; (ii) no SignBank variant, all of its BOBSL samples are filtered out; (iii) more than one SignBank variants, its BOBSL samples are assigned the SignBank variant that maximizes their cosine similarities.}

As mentioned in
\appendixref{Sec.~3.1}{\cref{subsec:data}} of the main paper,
the THMR used for this assignment is trained with random assignment (since initially there is no assignment).
But since we only use its motion encoder, we do not observe a big difference
in its assignment behavior if we use the second iteration of the THMR model,
trained on these better assigned texts. This may be due to a relatively
small search space within SignBank candidates.
However, note that for our evaluations, we use the latter THMR.

\section{Additional results}
\label{sec:app:experiments}

Here, we complement the results in the main paper
by
reporting additional metrics using motion-to-text retrieval for \dataset-Test (\cref{subsec:app:m2t}),
evaluating on SignBank as a test set (\cref{subsec:app:signbankresults}), and
visualizing LLM inputs and outputs (\cref{subsec:app:llm}).
We also provide additional qualitative visualizations (\cref{subsec:app:qualitative}).

\subsection{Motion-to-text retrieval metrics on \dataset-Test}
\label{subsec:app:m2t}
\begin{table*}
    \setlength{\tabcolsep}{3pt}
    \centering    
    \resizebox{1.0 \linewidth}{!}{
    \begin{tabu}{l|ccccc|ccccc}
        \toprule
        & \multicolumn{5}{c|}{Seen} & \multicolumn{5}{c}{Unseen} \\
        Input & R@1 $\uparrow$ & R@3 $\uparrow$ & FID $\downarrow$ & Div.$\leftarrow$ & MultiMod. $\uparrow$ & R@1 $\uparrow$ & R@3 $\uparrow$ & FID $\downarrow$ & Div.$\leftarrow$ & MultiMod. $\uparrow$ \\
        \midrule
        Ground truth & 22.58$_{\textcolor{white}{\pm 0.0}}$ $\mid$ 24.93$_{\textcolor{white}{\pm 0.0}}$  & 40.86$_{\textcolor{white}{\pm 0.0}}$ $\mid$ 42.80$_{\textcolor{white}{\pm 0.0}}$ & - & 0.94 $\mid$ 0.94 & - & 35.63$_{\textcolor{white}{\pm 0.0}}$ $\mid$ \textbf{34.48}$_{\textcolor{white}{\pm 0.0}}$ & \textbf{59.77}$_{\textcolor{white}{\pm 0.0}}$ $\mid$ \textbf{56.32}$_{\textcolor{white}{\pm 0.0}}$ & - & 0.93 $\mid$ 0.93 & - \\
        \midrule
        LLM(Phonology) & \textbf{52.80}$_{\pm 1.3}$ $\mid$ \textcolor{white}{0}8.21$_{\pm 0.2}$ & \textbf{74.06}$_{\pm 0.7}$ $\mid$ 16.10$_{\pm 0.4}$ & \textbf{0.15} $\mid$ 0.23 & \textbf{0.93} $\mid$ 0.90 & 0.09 $\mid$ \textbf{0.15} & \textbf{36.20}$_{\pm 0.6}$ $\mid$ 19.54$_{\pm 5.0}$ & 56.32$_{\pm 2.9}$ $\mid$ 39.66$_{\pm 0.6}$ & \textbf{0.38} $\mid$ 0.46 & 0.89 $\mid$ 0.89 & 0.10 $\mid$ 0.15 \\
        LLM(Phonology+HMS) & 51.63$_{\pm 0.8}$ $\mid$ \textbf{52.29}$_{\pm 0.9}$ & 72.51$_{\pm 0.4}$ $\mid$ \textbf{70.81}$_{\pm 0.9}$ & 0.16 $\mid$ \textbf{0.16} &   0.91 $\mid$ \textbf{0.91} & 0.09 $\mid$ 0.09 & 32.47$_{\pm 0.5}$ $\mid$ 30.74$_{\pm 2.9}$ & 51.72$_{\pm 1.4}$ $\mid$ 55.17$_{\pm 1.2}$ & 0.40 $\mid$ \textbf{0.40} & \textbf{0.90} $\mid$ 0.89 & \textbf{0.13} $\mid$ 0.15  \\
        \bottomrule
    \end{tabu}
    }
    \vspace{-0.2cm}
    \caption{\textbf{Motion-to-text metrics: } We complete the results of \appendixref{Tab.~2}{\cref{tab:input}} by providing the motion-to-\textit{text} retrieval scores for our LLM(Phonology) and LLM(Phonology+HMS) models,
    as well as the \dataset-Test ground truth motions.
    See text for comments.
    }
    \label{tab:app:input_m2t}
    \vspace{-0.3cm}
\end{table*}

\changed{For completeness, we repeat the last two rows of the \appendixref{Tab.~2}{\cref{tab:input}} of the main paper,
this time by reporting the motion-to-\textit{text} metrics from THMR (as opposed to the
motion-to-\textit{motion} metrics used in the main paper for \dataset).
The results in \cref{tab:app:input_m2t} %
support our previous conclusions drawn from the comparison of motion-to-motion scores. We also complement the metrics with diversity and multimodality scores for completeness, and include ground truth evaluation as a reference (first row).

On the unseen test set, performance of the ground-truth motion to retrieve text is overall better than that of generated motions (also because the descriptions are seen by the THMR as mentioned in \cref{subsec:app:thmr}).
On the seen partition, generations outperform ground-truth motions in terms of text retrieval capability using THMR.
A potential explanation is that the generations resemble the \textit{training} set, on which the THMR was optimized on.
This phenomenon is similar to the observations in \cite{petrovich2023tmr} (see Table A.2 of their paper).}

\subsection{Testing on SignBank motions}
\label{subsec:app:signbankresults}
\cref{tab:app:signbank} evaluates the two model variants
(corresponding to the last two rows of \appendixref{Tab.~2}{\cref{tab:input}} of the main paper)
by comparing the generated motions to ground truth SignBank motions. %
Motion-to-motion similarity suffers from a domain gap when comparing them with the shorter and quicker signs of BOBSL. Regardless, the conclusion that using HandMotionScript
benefits still holds.

\begin{table*}
	\setlength{\tabcolsep}{3pt}
	\centering    
	\resizebox{1\linewidth}{!}{
		\begin{tabular}{l|ccc|ccc}
			\toprule
			&  \multicolumn{3}{c|}{Seen}  &   \multicolumn{3}{c}{Unseen} \\
			Input & R@1 $\uparrow$ & R@03 $\uparrow$ & FID $\downarrow$ & R@1 $\uparrow$ & R@03 $\uparrow$ & FID $\downarrow$ \\
			\midrule
LLM(Phonology) & \textbf{17.56}$_{\pm 0.3}$ $\mid$ \textcolor{white}{0}5.54$_{\pm 0.4}$ & \textbf{30.78}$_{\pm 0.6}$ $\mid$ 11.74$_{\pm 0.4}$ & 0.36 $\mid$ \textbf{0.42} & \textbf{22.70}$_{\pm 1.7}$ $\mid$ 12.37$_{\pm 1.6}$ & 35.92$_{\pm 1.0}$ $\mid$ 24.45$_{\pm 2.0}$ & 0.57 $\mid$ \textbf{0.60} \\
LLM(Phonology+HMS) & 16.59$_{\pm 0.4}$ $\mid$ \textbf{17.73}$_{\pm 0.1}$ & 30.40$_{\pm 0.5}$ $\mid$ \textbf{32.55}$_{\pm 0.3}$ & 0.36 $\mid$ 0.36 & 20.98$_{\pm 2.5}$ $\mid$ \textbf{21.98}$_{\pm 2.8}$ & \textbf{41.67}$_{\pm 1.0}$ $\mid$ \textbf{34.06}$_{\pm 2.5}$ & \textbf{0.58} $\mid$ 0.57 \\
			\bottomrule
		\end{tabular}
	}
	\caption{\textbf{Transfer results on SignBank motions:}
		We report the two model variants from \appendixref{Tab.~2}{\cref{tab:input}} of the main paper.
		The models are trained on \dataset and tested on motions from \textit{SignBank}. We note a domain gap between co-articulated (fast) motions from \dataset compared to isolated (slow) motions from SignBank.
		The notation for the `$\mid$' separator is the same as before, i.e., `LLM(Phonology) $\mid$ LLM(Phonology+HMS)'.
	}
	\label{tab:app:signbank}
\end{table*}

\subsection{Qualitative results for the LLM}
\label{subsec:app:llm}
\begin{figure*}
	\centering
	\includegraphics[width=1\linewidth]{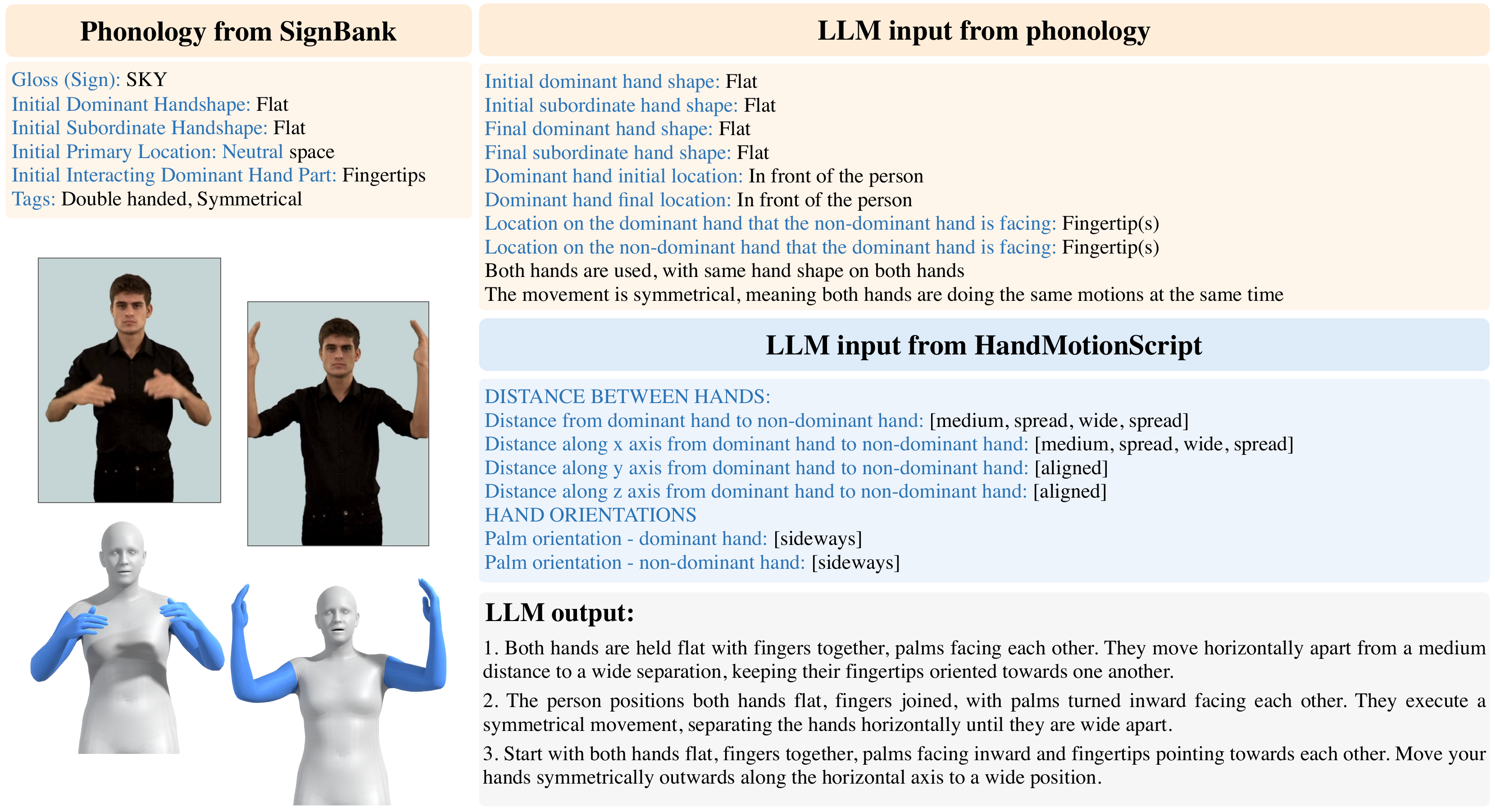}
	\vspace{-0.6cm}
	\caption{\textbf{Qualitative results for the LLM:}
		We show an example input-output for the LLM when converting
		the attributes from the SignBank phonology and from our HandMotionScript
		into a free-form language description.
		Note that we show the gloss (`SKY') for reference, but this information
		is not included in the LLM input.
		More examples are continued in \cref{fig:app:sb2}.
	}
	\label{fig:app:sb1}
\end{figure*}

\begin{figure*}
	\centering
	\includegraphics[width=.9\linewidth]{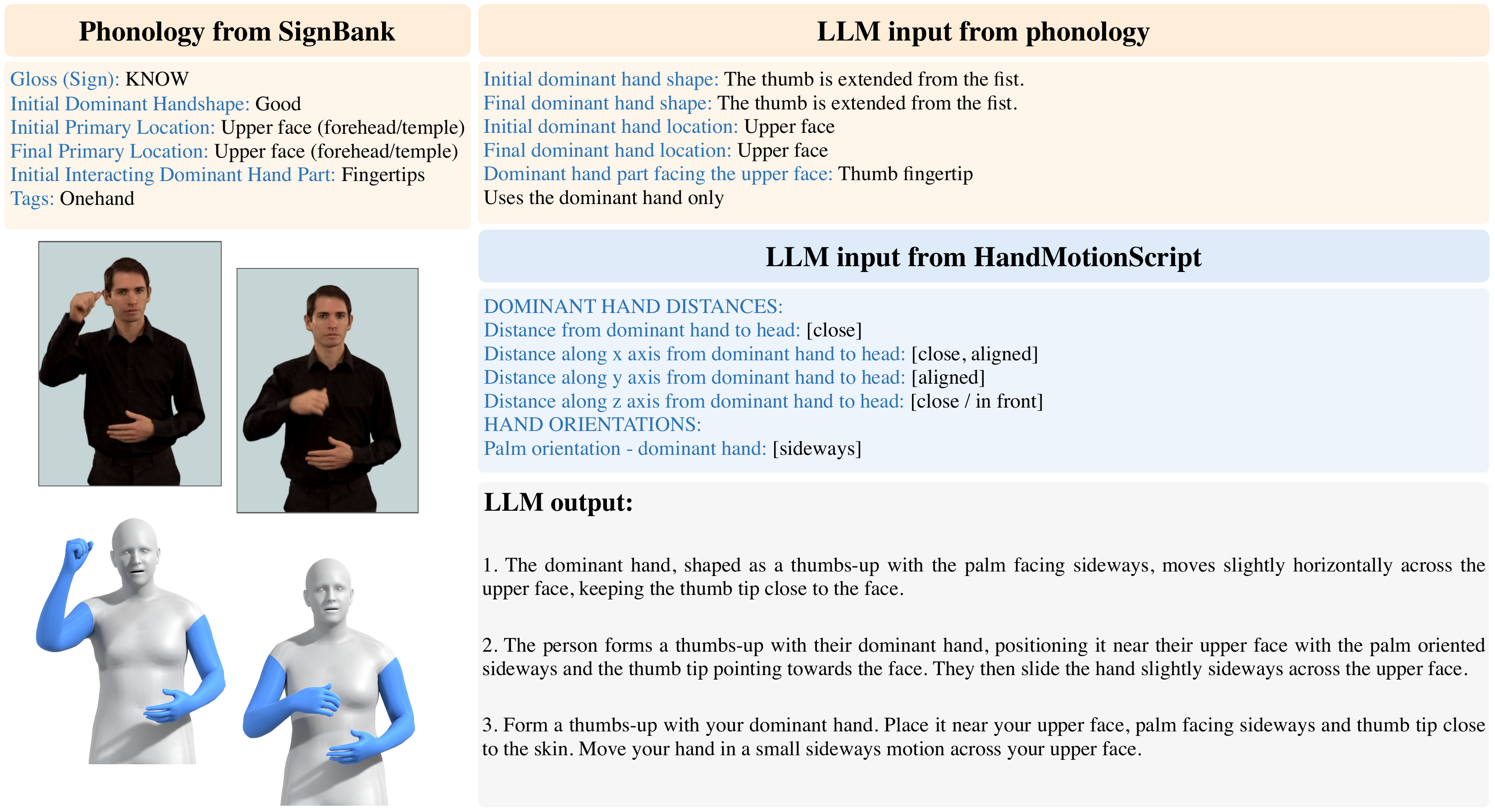}
	\includegraphics[width=.9\linewidth]{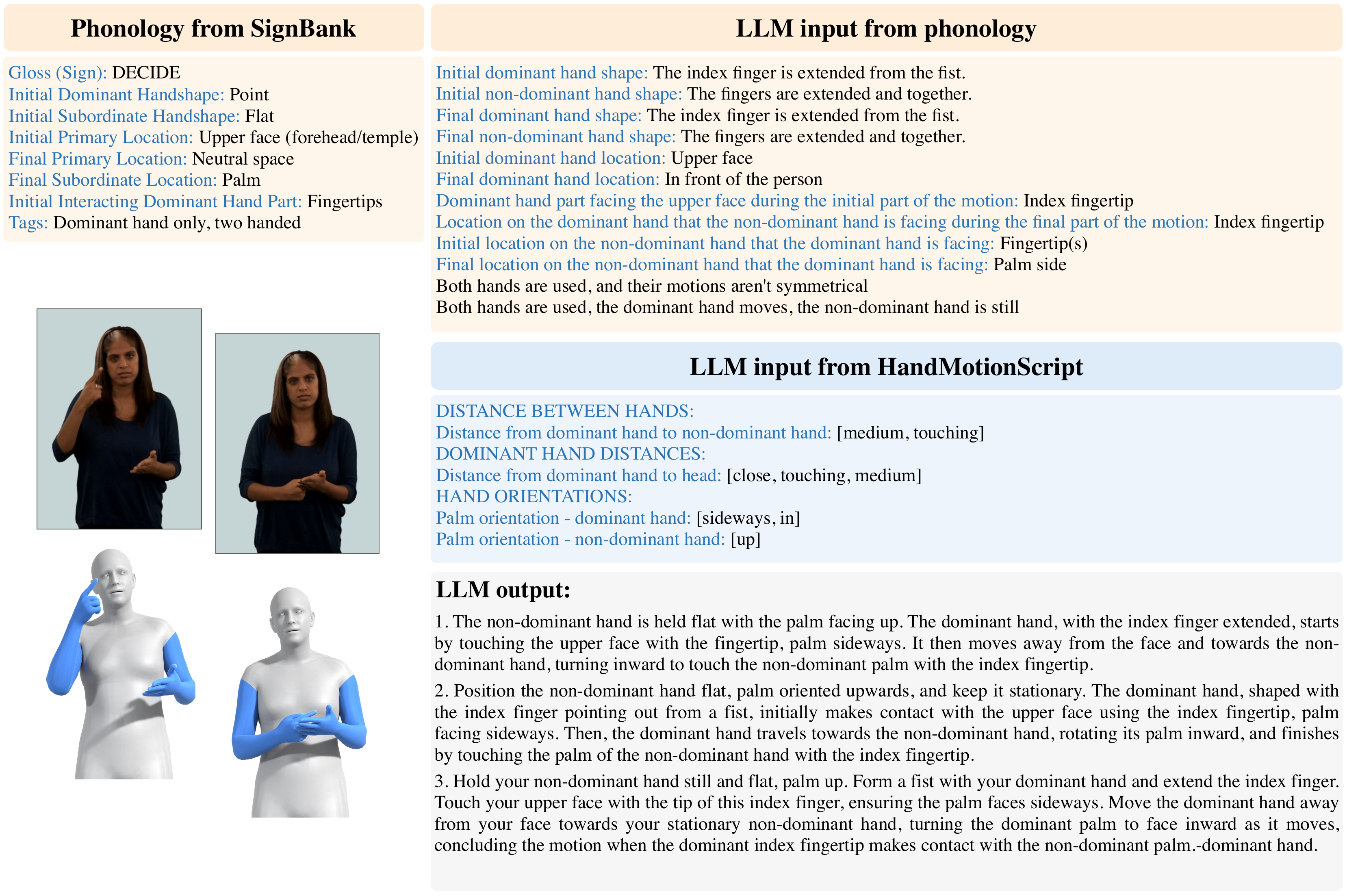}
	\vspace{-0.3cm}
	\caption{\textbf{\cref{fig:app:sb1} continued}
	}
	\label{fig:app:sb2}
\end{figure*}

In \cref{fig:app:sb1} and \cref{fig:app:sb2},
we provide example inputs and outputs for the conversion from
SignBank data to free-form textual descriptions.
We first map the linguistic attributes to descriptions
with our lookup table \changed{(as explained in
\cref{subsec:app:signbank})}, which are then input to the LLM
along with our HandMotionScript features.

\subsection{Additional qualitative visualizations}
\label{subsec:app:qualitative}

\begin{figure*}
	\centering
	\includegraphics[width=1\linewidth]{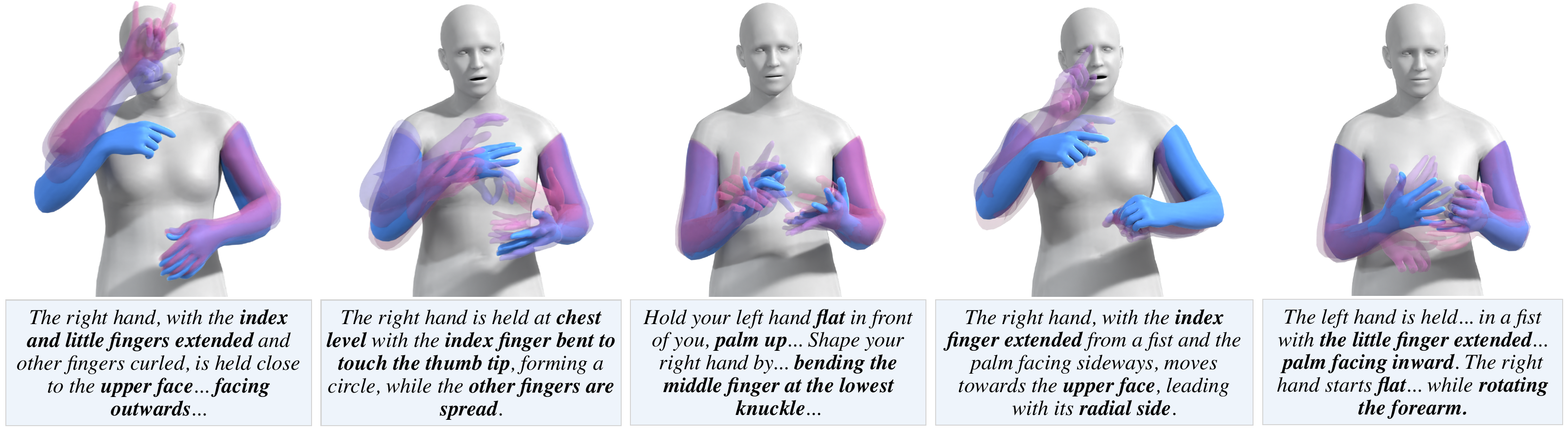}
	\vspace{-0.6cm}
	\caption{
		\textbf{Dynamic visualization for \appendixref{Fig.~1}{\cref{fig:teaser}} of the main paper:}
        We display the same examples as in \appendixref{Fig.~1}{\cref{fig:teaser}} with a dynamic style including 5 frames evenly sampled. The color coding denotes the temporal evolution, i.e., the last frame with blue, and the first frame with decreased transparency in pink.
    }
	\label{fig:app:teaser}
\end{figure*}

\begin{figure*}
	\centering
	\includegraphics[width=1\linewidth]{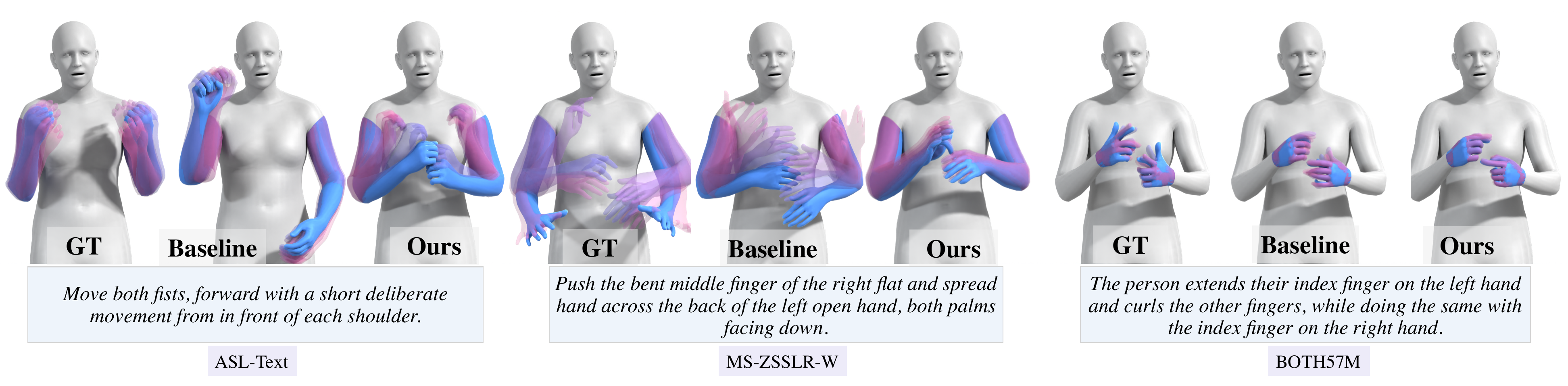}
	\vspace{-0.6cm}
	\caption{
		\textbf{Dynamic visualization for \appendixref{Fig.~4}{\cref{fig:qual_transfer}} of the main paper.}
	}
	\label{fig:app:qual_transfer}
\end{figure*}

\begin{figure*}
	\centering
	\includegraphics[width=1\linewidth]{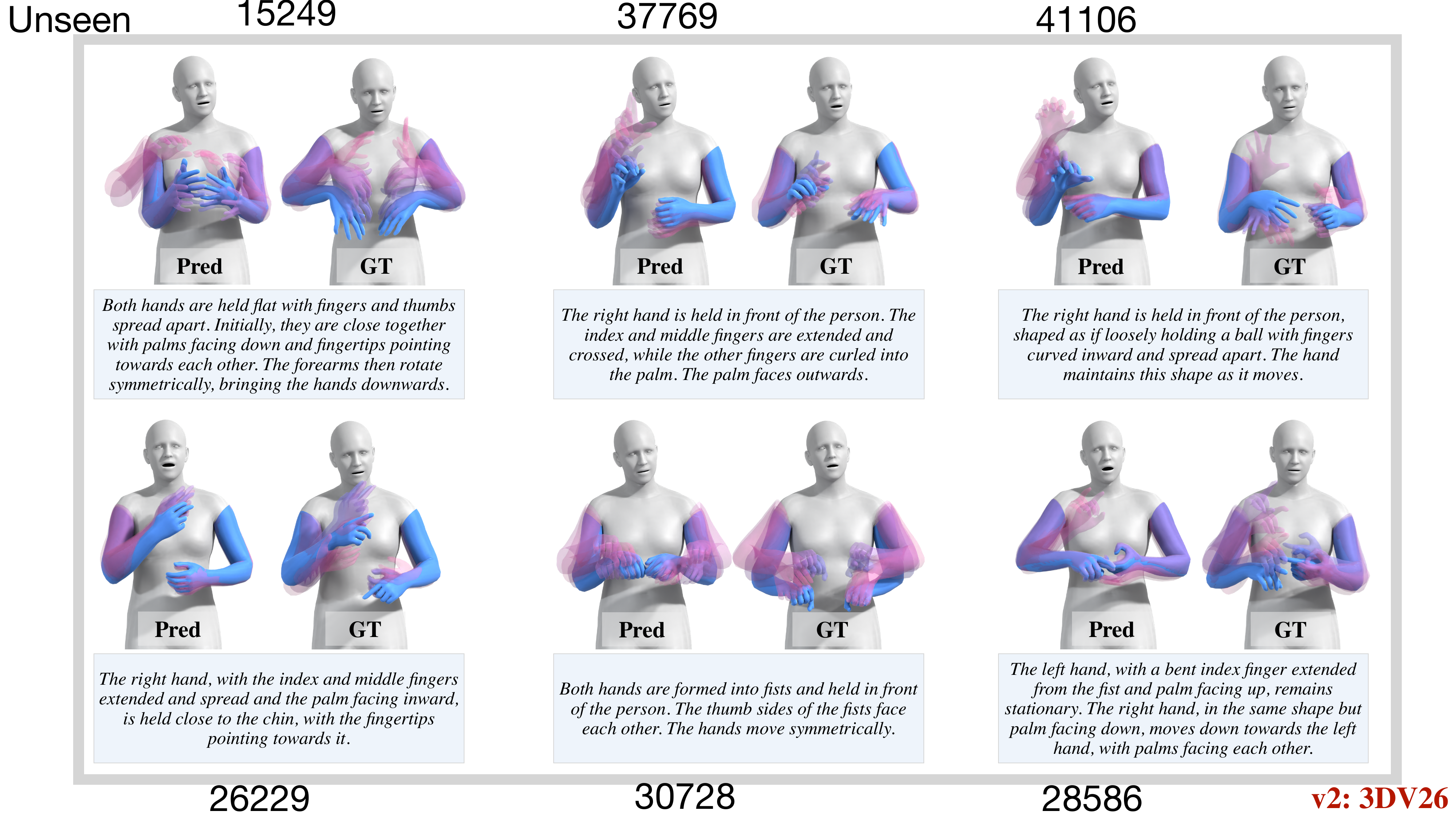}
	\vspace{-0.6cm}
	\caption{
	\textbf{Dynamic visualization for \appendixref{Fig.~5}{\cref{fig:qual}} of the main paper.}
    }
	\label{fig:app:qual}
\end{figure*}

\begin{figure*}
	\centering
	\includegraphics[width=1\linewidth]{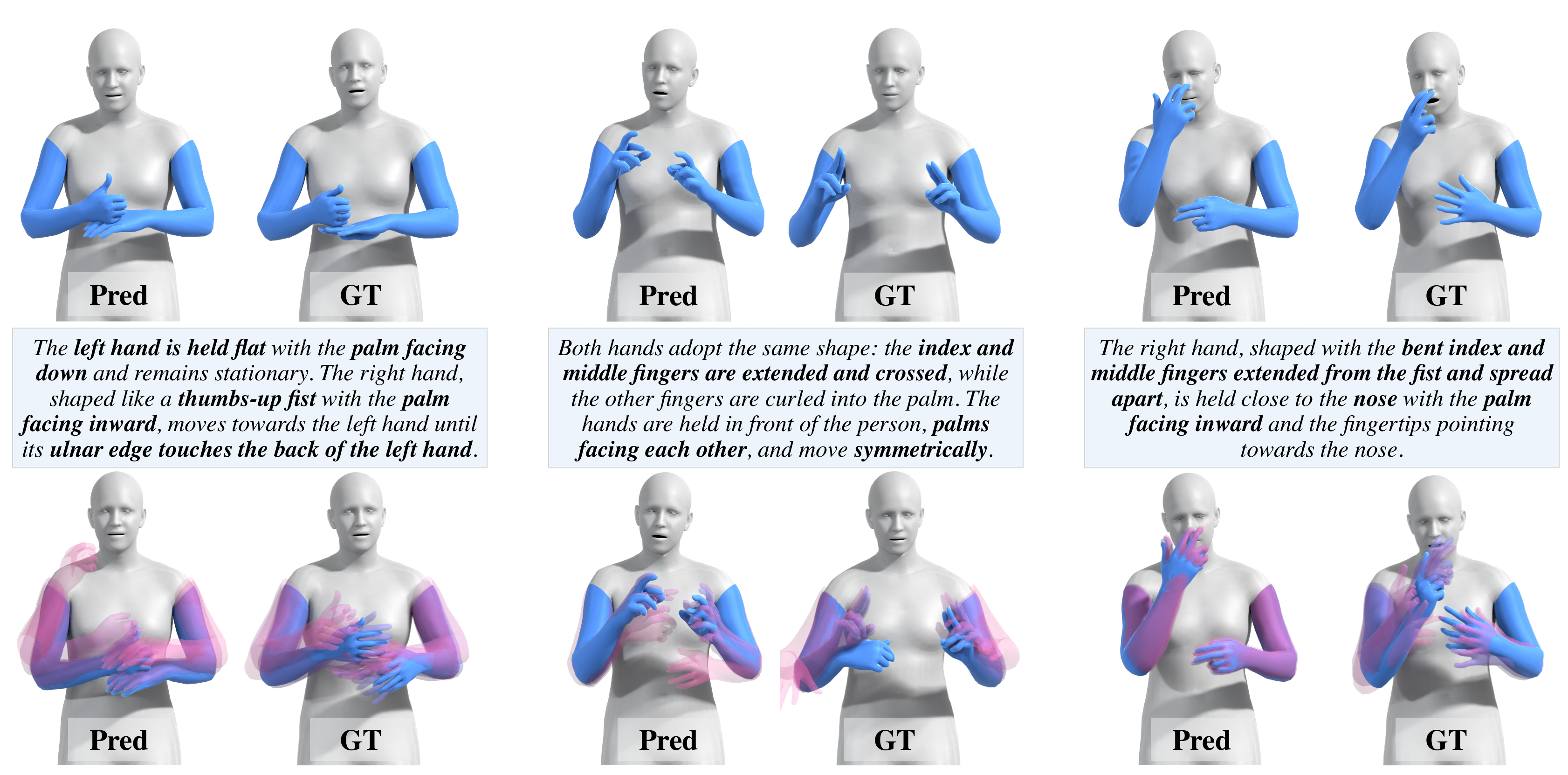}
	\vspace{-0.6cm}
	\caption{\textbf{Qualitative results on the seen test set of \dataset-Test:}
		We complement \appendixref{Fig.~5}{\cref{fig:qual} of the main paper (which shows results on the unseen test set),
		with results on the seen signs. We display both single-frame and dynamic visualizations for each example.}
	}
	\label{fig:app:qualseen}
\end{figure*}

We repeat the main paper figures in dynamic motion visualization in \cref{fig:app:teaser,fig:app:qual_transfer,fig:app:qual},
and show more results on \dataset-Test, this time on the \textit{seen} partition (\cref{fig:app:qualseen}).

\lstset{
  basicstyle=\footnotesize\ttfamily,  %
  breaklines=true,                  %
  breakatwhitespace=true,           %
  columns=fullflexible,             %
  keepspaces=true,                  %
  showstringspaces=false,           %
  frame=single,                     %
  framerule=0pt,
  rulecolor=\color{black},
  keywordstyle=\color{blue}\bfseries,
  stringstyle=\color{teal},
  commentstyle=\color{gray},
  numberstyle=\tiny\color{gray},
  identifierstyle=\color{black},
}

\begin{algorithm*}[htb!]
\caption{Prompt for LLM(Phonology + HMS) text generation.}
\label{alg:app:prompt1}
\begin{lstlisting}
You are an assistant whose role is to describe the hand motion for a sign in British Sign Language, given a list of information.
Use the information to describe the corresponding motion in 3 different ways.

In the ATTRIBUTES section, the given hand shapes concern the initial and final position. Having the information: 'Both hands are used, and their hand shapes are different', is not contradictory with the fact that their initial and final hand shapes are similar because they can be different in the middle of the motion.
In the DISTANCES sections:
- You are given temporal sequences of the distance between two limbs.
- The x-axis refers to the axis from left to right of the body, the y-axis refers to the axis from bottom to top, and the z-axis refers to the axis normal to the body and pointing away from it.
In the HAND ORIENTATIONS section:
- You are given temporal sequences of the palm orientations.

Use all the information in the ATTRIBUTES section unless you find that some points are contradictory.
Infer from the DISTANCES sections the directions and change of directions of the movements of the hands and include them in the description.
Infer from the HAND ORIENTATIONS section the main hand orientations of the hands during the movement and include them in the description.
Some points can be redundant, therefore summarize all the information into concise but exhaustive descriptions.
It is never the case that the dominant hand remains stationary. If you can't infer what it is doing from the information, just mention that it moves.

The 3 descriptions should describe the same motion using different wording.
Use a diversified vocabulary from one example to another.

Here are some examples. The comment section of the model answers, when present, is here for your reference as indications of the correct practices of the task, it is not to be included in your own output.

{
   {
       "role": "user",
       "content": "Use the information to describe the corresponding motion. The comment section of the model answers, when present, is here for your reference as indications of the correct practices of the task, it is not to be included in your own output.
       ATTRIBUTES:
       - Initial dominant hand shape: Pointing
       - Initial non-dominant hand shape: Pointing
       - Final dominant hand shape: Pointing
       - Final non-dominant hand shape: Pointing
       - Initial sign location: In front of the person
       - Final sign location: In front of the person
       - Location on the dominant hand that the non-dominant hand is facing: Broad side
       - Location on the non-dominant hand that the dominant hand is facing: Broad side
       - Both hands are used, with same hand shape on both hands
       - The movement is symmetrical, meaning both hands are doing the same motions at the same time
       DISTANCE BETWEEN HANDS:
       - Distance from dominant hand to non-dominant hand: [spread, close]
       - Distance along x axis from dominant hand to non-dominant hand: [spread, close]
       - Distance along y axis from dominant hand to non-dominant hand: [aligned]
       - Distance along z axis from dominant hand to non-dominant hand: [aligned]
       HAND ORIENTATIONS:
       - Palm orientation - dominant hand: [sideways]
       - Palm orientation - non-dominant hand: [sideways]"
   },
   {
       "role": "model",
       "content": {
           "Description 1": "Both hands, pointing and facing sideways, palm facing each other, and are moving toward each other horizontally.",
           "Description 2": " The person positions both hands so that the index fingers are extended and the hands are pointing sideways, with the palms turned inward to face each other. Then, they move their hands horizontally toward one another.",

\end{lstlisting}
\end{algorithm*}

\begin{algorithm*}[htb!]
\caption{Prompt for LLM(Phonology + HMS) text generation. (Algorithm 2 continued)}
\label{alg:app:prompt2}
\begin{lstlisting}
           "Description 3": " Extend both index fingers and place the hands with the palm facing each other. Then move your hands horizontally symmetrically toward one another.",
           "Comment": "When the hands are facing sideways, it is always towards one another, not towards the outside. Besides, the hands are never crossed when the location is in front of the person.
           They face each other on their broad side, meaning not on one of their edges. Frome these 3 points you can deduce that their palms are facing each other.
           When you can deduce this kind information, add it to at least one of the descriptions.
           The horizontal aspect of the motion is deduced from the fact that the only direction with a change in the distance between the hands is the x axis."
       }
   }
},
{
   {
       "role": "user",
       "content": "Use the information to describe the corresponding motion. The comment section of the model answers, when present, is here for your reference as indications of the correct practices of the task, it is not to be included in your own output.
       ATTRIBUTES:
       - Initial dominant hand shape: The index forms a hook. The thumb is extended and stuck against the index finger. The other fingers are curled on the palm.
       - Final dominant hand shape: The index forms a hook. The thumb is extended and stuck against the index finger. The other fingers are curled on the palm.
       - Dominant hand location: Chest level
       - Dominant hand part facing the chest: Hand edge on the thumb side
       - Uses the dominant hand only
       DOMINANT HAND DISTANCES:
       - Distance from dominant hand to torso: [close]
       - Distance along x axis from dominant hand to torso: [aligned]
       - Distance along y axis from dominant hand to torso: [aligned]
       - Distance along z axis from dominant hand to torso: [close / in front]
       HAND ORIENTATIONS:
       - Palm orientation - dominant hand: [sideways]"
   },
   {
       "role": "model",
       "content": {
           "Description 1": "The dominant hand is held shaped as a hook, with the index fingertip touching the thumb. It moves close to the chest, with its radial side facing the chest.",
           "Description 2": "The person holds his dominant hand in a hook shape, bringing the index fingertip to meet the thumb. The hand is positioned near the chest, with the thumb-side turned toward the chest.",
           "Description 3": "Form your dominant hand into a hook. Extend the thumb against the index finger, and make their fingertips touch. Bring the hand close to your chest, keeping the thumb-side facing toward your chest."
       }
   }
},
{
   {
       "role": "user",
       "content": "Use the information to describe the corresponding motion. The comment section of the model answers, when present, is here for your reference as indications of the correct practices of the task, it is not to be included in your own output.
       ATTRIBUTES:
       - Initial dominant hand shape: The fingers are extended and together.
       - Initial non-dominant hand shape: The fingers are extended and together.
       - Final dominant hand shape: Round
       - Final non-dominant hand shape: Round
       - Sign location: Chest level
       - Dominant hand part facing the chest: Fingertip(s)
       - Subordinate hand part facing the chest: Fingertip(s)
       - Both hands are used, with same hand shape on both hands
       - Hand shapes change during the sign
       - The movement is symmetrical, meaning both hands are doing the same motions at the same time
\end{lstlisting}
\end{algorithm*}

\begin{algorithm*}[htb!]
\caption{Prompt for LLM(Phonology + HMS) text generation. (Algorithm 3 continued)}
\label{alg:app:prompt3}
\begin{lstlisting}
       DOMINANT HAND DISTANCES:
       - Distance from dominant hand to torso: [medium, close]
       NON-DOMINANT HAND DISTANCES:
       - Distance from non-dominant hand to torso: [medium, close]
       HAND ORIENTATIONS:
       - Palm orientation - dominant hand: [down]
       - Palm orientation - non-dominant hand: [down]"
   },
   {
       "role": "model",
       "content": {
           "Description 1": "The hands start flat with fingers together. They move with the palm facing down toward the chest, while closing into a round shape.",
           "Description 2": "The person begins with both hands flat and fingers held together. As the hands move inward toward the chest with palms facing downward, they gradually form a rounded shape.",
           "Description 3": "Start with your hands flat and fingers together. Move them toward your chest with palms facing down, and as you do, curve the hands into a round shape.",
           "Comment": ""The fact that the sign location is the chest doesn't mean that the hands are at the chest during the whole sign.
           It means that the hands are interacting with the chest, or positioned near the chest, at least at some point of the motion.
           Here, from the distance information, you can deduce that the motion consists in the hands coming to the chest."
       },
   }
},
{
   {
       "role": "user",
       "content": "Use the information to describe the corresponding motion. The comment section of the model answers, when present, is here for your reference as indications of the correct practices of the task, it is not to be included in your own output.
       ATTRIBUTES:
       - Initial dominant hand shape: Pointing
       - Initial non-dominant hand shape: The fingers are extended and together.
       - Final dominant hand shape: The hand is held flat with the fingers held together.
       - Final non-dominant hand shape: The fingers are extended and together.
       - Initial dominant hand location: Upper face
       - Final dominant hand location: In front of the person
       - Dominant hand part facing the upper face during the initial part of the motion: Index fingertip
       - Location on the dominant hand that the non-dominant hand is facing during the final part of the motion: Back side
       - Initial subordinate hand part facing the upper face: Fingertip(s)
       - Final subordinate hand part facing the upper face: Palm side
       - Forearm rotation
       - Both hands are used, and their motions aren't symmetrical
       - Hand shapes change during the sign
       DOMINANT HAND DISTANCES:
       - Distance from dominant hand to head: [close, touching, close, medium]
       DISTANCE BETWEEN HANDS:
       - Distance from dominant hand to non-dominant hand: [spread, medium, close]
       HAND ORIENTATIONS:
       - Palm orientation - dominant hand: [sideways, in]
       - Palm orientation - non-dominant hand: [up]
   },
   {
       "role": "model",
       "content": {
           "Description 1": "The dominant touches the head with its extended index finger and the palm facing sideways. Then the hands move toward each other, both held flat, with the non-dominant hand turned upward. The motion ends with the back of the dominant hand turned towards the palm of the non-dominant hand.",
           "Description 2": "The individual touches her head with the pointed index finger of her dominant hand, with the palm facing sideways. Then, she moves both hands toward each other, holding them flat, with the dominant hand turning inward and the non-dominant hand turned upward, to end the motion with the back of the dominant hand facing the palm of the non-dominant hand.",

\end{lstlisting}
\end{algorithm*}

\begin{algorithm*}[htb!]
\caption{Prompt for LLM(Phonology + HMS) text generation. (Algorithm 4 continued)}
\label{alg:app:prompt4}
\begin{lstlisting}

           "Description 3": "Move your dominant index finger to your head. Then, bring both hands toward each other, keeping them flat, with the non-dominant hand palm facing up. Ensure to end the movement with the back of your dominant hand facing the palm of your non-dominant hand.",
           "Comment": "The distance sequence from the dominant hand to the head indicates that the motion starts near the head and then farther away from it.
           On the contrary, the distance sequence between the dominant hand and the non-dominant hand indicates that the dominant hand starts far away from the non-dominant hand and gets closer to it during the motion.
           From these 2 points you deduce that the dominant hand starts by touching the head, then that both hands move toward each other.
           All the points don't need to be included in each of the descriptions, especially when they seem redundant or necessary. For instance, if the dominant hand goes from the head to the non-dominant, that the non-dominant hand palm is facing up, and that the dominant hand meets the non-dominant hand palm with its back, then the dominant hand will necessarily face in during this motion."
       }
   }
},
{
   {
       "role": "user",
       "content": "Use the information to describe the corresponding motion. The comment section of the model answers, when present, is here for your reference as indications of the correct practices of the task, it is not to be included in your own output.
       ATTRIBUTES:
       - Initial dominant hand shape: The fingers are held together and tightly bent into a fist.
       - Initial non-dominant hand shape: The fingers are held together and tightly bent into a fist.
       - Final dominant hand shape: The fingers are held together and tightly bent into a fist.
       - Final non-dominant hand shape: The fingers are held together and tightly bent into a fist.
       - Initial sign location: Shoulder
       - Final sign location: Chest level
       - Dominant hand part facing the shoulder during the initial part of the motion: Dorsal side
       - Dominant hand part facing the chest during the final part of the motion: Dorsal side
       - Initial subordinate hand part facing the shoulder: Dorsal side
       - Final subordinate hand part facing the shoulder: Dorsal side
       - Both hands are used, with same hand shape on both hands
       - The movement is symmetrical, meaning both hands are doing the same motions at the same time
       DOMINANT HAND DISTANCES:
       - Distance from dominant hand to non-dominant shoulder: [medium]
       - Distance from dominant hand to dominant shoulder: [close]
       - Distance from dominant hand to torso: [close]
       NON-DOMINANT HAND DISTANCES:
       - Distance from non-dominant hand to dominant shoulder: [medium]
       - Distance from non-dominant hand to non-dominant shoulder: [close]
       - Distance from non-dominant hand to torso: [close]
       HAND ORIENTATIONS:
       - Palm orientation - dominant hand: [in]
       - Palm orientation - non-dominant hand: [in]"
   },
   {
       "role": "model",
       "content": {
           "Description 1": "Both hands form a fist, and are turned inward. They move near their respective shoulders.",
           "Description 2": "The person bends all his fingers into a fist, turning them inward with the dorsal side of the hand facing the shoulders. He then moves his fists close to their respective shoulders.",
           "Description 3": "Form a fist with both hands, turning them inward. Move your fists close to your shoulders.",
           "Comment": "When the shoulders are mentioned in the location, you need to deduce which shoulder each hand is located at, usually by checking which shoulder each hand is the closest to - here their respective shoulders. The hand part facing the shoulder sections and the palm orientation sections are redundant. In this case, alternate the vocabulary between descriptions."
       }
   }
},

\end{lstlisting}
\end{algorithm*}

\begin{algorithm*}[htb!]
\caption{Prompt for LLM(Phonology + HMS) text generation. (Algorithm 5 continued)}
\label{alg:app:prompt5}
\begin{lstlisting}

{
   {
       "role": "user",
       "content": "Use the information to describe the corresponding motion. The comment section of the model answers, when present, is here for your reference as indications of the correct practices of the task, it is not to be included in your own output.
       ATTRIBUTES:
       - Initial dominant hand shape: The little finger is extended, all the other fingers are curled on the palm
       - Initial non-dominant hand shape: The hand is held flat with the fingers held together.
       - Final dominant hand shape: The little finger is extended, all the other fingers are curled on the palm
       - Final non-dominant hand shape: The hand is held flat with the fingers held together.
       - Initial dominant hand location: In front of the person
       - Final dominant hand location: In front of the person
       - Location on the dominant hand that the non-dominant hand is facing: Ulnar side
       - Location on the non-dominant hand that the dominant hand is facing: Palm side
       - Forearm rotation
       - Both hands are used, and their motions aren't symmetrical
       - Both hands are used, the dominant hand moves, the non-dominant hand is still
       DISTANCE BETWEEN HANDS:
       - Distance from dominant hand to non-dominant hand: [close, medium]
       HAND ORIENTATIONS:
       - Palm orientation - dominant hand: [up, out]
       - Palm orientation - non-dominant hand: [up]"
   },
   {
       "role": "model",
       "content": {
           "Description 1": "The non-dominant hand is held flat with the palm facing up. The dominant hand, starting with the little finger extended from the fist, palm facing up and its ulnar side on the non-dominant hand, moves away from the non-dominant hand.",
           "Description 2": "The person holds their non-dominant hand flat with the palm facing upward. Their dominant hand begins in a fist with only the little finger extended, palm also facing up, and the little finger edge of the hand resting on the non-dominant palm. They then move the dominant hand away from the non-dominant hand.",
           "Description 3": "Hold the non-dominant hand flat with the palm facing upward. Form a fist with the dominant hand, extending only the little finger, and position it palm up with the ulnar edge resting on the non-dominant palm. Then move the dominant hand away from the non-dominant hand."
       }
   }
}
\end{lstlisting}
\end{algorithm*}

\end{document}